\pdfoutput=1

\documentclass[11pt]{article}

\usepackage[final]{acl}

\usepackage{times}
\usepackage{latexsym}

\usepackage[T1]{fontenc}

\usepackage[utf8]{inputenc}

\usepackage{microtype}

\usepackage{inconsolata}

\usepackage{graphicx}

\usepackage{amsmath}
\usepackage{subcaption}

\usepackage{multirow}

\title{Cluster-Norm for Unsupervised Probing of Knowledge}

\author{
 \textbf{Walter Laurito\textsuperscript{1,2, *}},
 \textbf{Sharan Maiya\textsuperscript{2,3, *}},
 \textbf{Grégoire Dhimoïla\textsuperscript{4,2, *}},  \\
 \textbf{Ho Wan Yeung\textsuperscript{5}}, 
 \textbf{Kaarel Hänni\textsuperscript{6}}
\\
\\
 \textsuperscript{1}FZI,
 \textsuperscript{2}Cadenza Labs,
 \textsuperscript{3}University of Cambridge,
 \textsuperscript{4}ENS Paris-Saclay,
 \textsuperscript{5}UC Berkeley,
 \textsuperscript{6}Caltech
\\
 \small{
   \textbf{Correspondence:} \href{mailto:laurito@fzi.de}{laurito@fzi.de}
 }
}

\begin{document}
\maketitle
\def\thefootnote{*}\footnotetext{These authors contributed equally to this work.}\def\thefootnote{\arabic{footnote}}

\begin{abstract}
The deployment of language models brings challenges in generating reliable information, especially when these models are fine-tuned using human preferences. To extract encoded knowledge without (potentially) biased human labels, unsupervised probing techniques like Contrast-Consistent Search (CCS) have been developed \citep{burns2022discovering}. However, salient but unrelated features in a given dataset can mislead these probes \citep{farquhar2023challenges}. Addressing this, we propose a cluster normalization method to minimize the impact of such features by clustering and normalizing activations of contrast pairs before applying unsupervised probing techniques. While this approach does not address the issue of differentiating between knowledge in general and simulated knowledge—a major issue in the literature of latent knowledge elicitation \citep{christianoELK2021}—it significantly improves the ability of unsupervised probes to identify the intended knowledge amidst distractions.\footnote{The code for this work is available at \url{https://github.com/Cadenza-Labs/cluster-normalization}.}.
\end{abstract}

\section{Introduction}
\begin{figure*}[htpb]
    \centering
    \includegraphics[width=.9\linewidth]{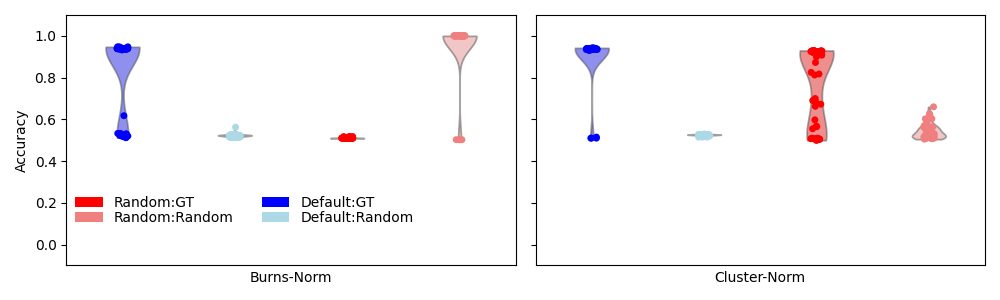}
\caption{Under the standard CCS approach (left; Burns-Norm), a modified prompt including distracting random words (red) causes all CCS probes to achieve random accuracy against ground truth labels (GT), and high accuracy against these random word labels (random). When using our approach of cluster normalization (right; Cluster-Norm), the average accuracy of CCS probes for the desired feature increases significantly.}
\label{fig:exp1_ccs}
\end{figure*}
The deployment of language models for practical applications introduces novel challenges, including the potential creation of untrustworthy or incorrect text \citep{weidinger2021ethical, park2023ai, evans2021truthful, hendrycks2021unsolved}. Specifically, models that are fine-tuned using human preferences may amplify existing human biases or generate persuasive yet deceptive outputs \citep{perez2022discovering}. 

Empirical evidence suggests that simulated internal beliefs or \textit{knowledge} can be extracted from language model activations \citep{li2022emergent, gurnee2023language, azaria2023internal, bubeck2023sparks}. Supervised probing methods can be employed to extract this knowledge \citep{alain2016understanding, marks2023geometry} but such methods require labels, which in some domains may not be readily provided due to human biases or because humans simply do not know the correct label. It may even be critical to avoid the use of human labels to differentiate between a model's true knowledge and its representation of human knowledge. Motivated by these ideas, unsupervised probing techniques like Contrast-Consistent Search (CCS) have been developed to extract the knowledge embedded in a language model without the need for ground truth labels \citep{zou2023representation, burns2022discovering}. 

\citet{farquhar2023challenges} outline current limitations of these approaches, demonstrating that these unsupervised probes tend to identify the most salient binary feature, which may not always correspond to the specific knowledge feature we seek. For example, in one experiment, one of a pair of distracting random words is added to each prompt in a text dataset. After training, unsupervised CCS probes often function as classifiers for these random words, rather than the intended knowledge feature. In practice, there may be numerous salient features of which we are unaware, which can divert an unsupervised probe from identifying the target feature, regardless of whether they are correlated or uncorrelated with the target. 

To tackle this issue, we propose a cluster normalization method. 
Our method starts by following the usual initial approach of unsupervised probing of harvesting contrast pair activations. We then cluster similar activations and normalize them separately, thereby eliminating the effect of distracting salient features. We can then apply any unsupervised probing method, such as CCS or CRC-TPC \citep{burns2022discovering}, to train a probe on these normalized activations.
It is of course crucial to ensure this approach does not inadvertently eliminate the knowledge feature itself. To prevent this, we utilize contrast pairs, performing the clustering on the average embedding of each pair. Further details on contrast pairs are provided in Section \ref{ccs}.

Probes trained with the original CCS approach achieve an average accuracy of approximately 0.5 on prompt datasets with distracting random word features \citep{farquhar2023challenges}. In contrast, our clustering method significantly improves this average accuracy to about 0.77, and to 0.81 for CRC-TPC when tested with Mistral-7B. Details of our method are described in Section \ref{sec:method}. 

\section{Background}

\subsection{Contrast-Consistent Search (CCS)}
\label{ccs}

Contrast-Consistent Search (CCS), as described by \citet{burns2022discovering}, locates a direction in activation space using a perceptron that adheres to logical consistency principles. This is achieved through a loss function designed to ensure that probabilities for a question-answer pair and its negated counterpart — a contrast pair — are complementary. This loss function is optimized in an unsupervised manner, and in doing so CCS extracts the latent knowledge within large language models to answer binary questions.

At first, a language model $\mathcal{M}$ processes a dataset of textual contrast pairs ${(x_i^+, x_i^-)}_{i=1}^{n}$, generating contextualized embeddings ${(\mathcal{M}(x_i^+), \mathcal{M}(x_i^-))}$. Following this, a linear probe \citep{alain2016understanding} is trained to calculate from these embeddings the probabilities ${p^+}$ and ${p^-}$, whether ${x_i^+}$ or ${x_i^-}$ is true, respectively. The objective function used to train this probe is given by a sum of two terms:

{\small
\begin{align*}
& \mathcal{L}_{\text {CCS }}=\sum_{i=1}^N \mathcal{L}_{\text {consistency }}+\mathcal{L}_{\text {confidence }} \\
& \mathcal{L}_{\text {consistency }}=\left[p\left(x_i^{+}\right)-\left(1-p\left(x_i^{-}\right)\right)\right]^2 \\
& \mathcal{L}_{\text {confidence }}=\min \left\{p\left(x_i^{+}\right), p\left(x_i^{-}\right)\right\}^2 .
\end{align*}
}

The first term, \( L_{\text{consistency}} \), is motivated by the idea that the probabilities of a statement and its negation should sum to one. This ensures logical consistency. The second term, \( L_{\text{confidence}} \), is designed to maximize the information extracted by the probe, penalizing cases where the probabilities for both true and false are the same, at \( p(x^{+}) = p(x^{-}) = 0.5 \). Thus, this term encourages the probe to be more certain in its outputs.

Intuitively, there are at least two possible directions (features) satisfying this loss. The first is the knowledge direction we seek, $\vec{F}_{\top/\bot}$, and the second is the syntactical difference between positive and negative prompt templates, $\vec{F}_{\pm}$. To remove this latter undesired feature, \citet{burns2022discovering} proceed as follows. Before training an unsupervised probe, contrast pair activations are first normalized: $\widetilde{\mathcal{M}}(x_i^+) = \frac{\mathcal{M}(x_i^+) - \mu^+}{\sigma^+}$, with $\mu^+$ and $\sigma^+$ the mean and standard deviation of the activations of all positive examples in each contrast pair; the same normalization procedure is followed for negative examples, and the unsupervised probe is trained on these normalized $\widetilde{\mathcal{M}}(x_i^\pm)$. In this way, CCS removes the most salient feature of the contrast pair differences: the syntactical difference direction $\vec{F}_{\pm}$. However, as \citet{farquhar2023challenges} show, the second-most salient feature may not necessarily be the desired knowledge - an implicit assumption of the original CCS method. In this work, we take advantage of normalization to remove other undesired salient features by including a clustering step.

\subsection{Contrastive Representation Clustering}
\label{sec:CRC}
As an alternative to CCS, the method of \textit{Contrastive Representation Clustering via Top Principal Component} (CRC-TPC) \citep{burns2022discovering} separates the normalized contrast pair differences $\{\widetilde{\mathcal{M}}(x_i^+) - \widetilde{\mathcal{M}}(x_i^-)\}$ based on projections onto their top principal component, i.e., the singular vector associated with the highest singular value, or the direction with the highest variance. This is again motivated by the intuition that the most salient contrastive feature after $\vec{F}_{\pm}$ - removed by normalization - should be the knowledge feature $\vec{F}_{\top/\bot}$.

\subsection{Theoretical Background}
\label{sec:theoretical_background}

A \textit{salient feature} is a direction with high variance in the data. We are interested in salient features in the contrast pair differences $\widetilde{\mathcal{M}}(x_i^+) - \widetilde{\mathcal{M}}(x_i^-)$, and we refer to these as \textit{contrastive features}.

In this section, we explain (1) why undesired salient contrastive features can mislead unsupervised probes, and (2) how contrastive features can be induced by non-contrastive ones. The mechanisms of the latter point are illustrated through an example.

We shall first examine why there is a close link between the CCS loss described in Section \ref{ccs} and the idea of saliency i.e., variance. Contrastive features will naturally achieve a low CCS loss. To see this, consider the variance of contrast pair differences projected along the feature direction of a given feature $\vec{F}$:

{\small
\begin{align*}
    X &:= \vec{F}^T \cdot \widetilde{\mathcal{M}}(x_i^+),~Y:= \vec{F}^T \cdot \widetilde{\mathcal{M}}(x_i^-).\\
    Var(&X - Y) = \underbrace{E(X^2) + E(Y^2)}_{Confidence} \underbrace{- 2\cdot E(X \cdot Y)}_{Consistency} \\
    &\underbrace{-(E(X)^2 + E(Y)^2 - 2 \cdot E(X) \cdot E(Y))}_{=~0}
\end{align*}
}

In this expanded form, we see that the variance of contrast pair differences in the direction $\vec{F}$ captures,
    \begin{itemize}
    \item \textbf{confidence}, with $E(X^2) + E(Y^2)$ higher if the magnitude of the projection along $\vec{F}$ in either element of a pair is high, 
    \item and \textbf{consistency}, with $- 2\cdot E(X \cdot Y)
    > 0$ if the projections of a contrast pair along $\vec{F}$ have opposing sign. This also increases with the magnitude of these projections.
\end{itemize}

Note that the term $-(E(X)^2 + E(Y)^2 - 2 \cdot E(X) \cdot E(Y))$ equals zero under the set-up of CCS, as the normalization step described above results in $E(\widetilde{\mathcal{M}}(x_i^\pm)) = 0$, therefore $E(X) = E(Y) = 0$.

Due to this link between confidence, consistency, and contrastive saliency, a probe trained using the CCS loss will favor learning salient contrastive features. Otherwise, the projections of a contrast pair onto a feature will be small in difference or equal, failing to satisfy at least the consistency condition.

As mentioned in Section \ref{ccs}, we can describe two features which intuitively will satisfy the CCS loss:
\begin{itemize}
    \item $\vec{F}_{\pm}:=\vec{F}_{+}~-~\vec{F}_{-}$, the syntactical difference between contrast pairs due to the appending of positive and negative tokens, removed by normalization in the original CCS method,
    \item $\vec{F}_{\top/\bot}:=\vec{F}_{\top}~-~\vec{F}_{\bot}$, the knowledge feature we seek.
\end{itemize}

Under our definition, undesired distracting features such as proxies for knowledge or random words (as in \citet{farquhar2023challenges}) should not be contrastive features: their contrast pair projections should be equal in both examples of each pair and thus should be ignored by a CCS probe.
It is however the case that these non-contrastive features can still mislead unsupervised probes by \textit{inducing} undesired contrastive features. We describe the mechanism through which this occurs with the following example:

Let $f$ be some binary function, say the $XOR$ function on the presence of features, and $\vec{F}_1$, $\vec{F}_2$ be any two features. Suppose that a model represents the feature $f(\vec{F}_1, \vec{F}_2)$ as its own direction $\vec{F}_{f(\vec{F}_1, \vec{F}_2)}$, orthogonal to $\vec{F}_{1,2}$\footnote{Evidence of such behavior has been observed with $f = XOR$ and any two features, as shown in \cite{marks2024Xor}.}. Now, fix two features $\vec{F}_1$ and $\vec{F}_2$ and assume without loss of generality that exactly one of them appears in each pair with probability~$\frac{1}{2}$.

We can now write the contrast pair differences as:

{\small
\begin{align*}
\mathcal{M}(x_i^+) - &\mathcal{M}(x_i^-) = \underbrace{\vec{F}_{+} - \vec{F}_{-}}_{\vec{F}_{\pm}} + \underbrace{\vec{F}_{f(\vec{F}_{+}, \vec{F}_j)} - \vec{F}_{f(\vec{F}_{-}, \vec{F}_j)}}_{\Delta_{\vec{F}_j}^{\pm}} \\
&\pm (\underbrace{\vec{F}_{\top} - \vec{F}_{\bot}}_{\vec{F}_{\top/\bot}}) \pm (\underbrace{\vec{F}_{f(\vec{F}_{\top}, \vec{F}_j)} - \vec{F}_{f(\vec{F}_{\bot}, \vec{F}_j)}}_{\Delta_{\vec{F}_j}^{\top}})
\end{align*}
}
for $j \in \{1, 2\}$.

The expected value of these contrast pair differences over our dataset is:
{\small
\begin{equation}
\label{eq:mean}
    E(\mathcal{M}(x_i^+) - \mathcal{M}(x_i^-)) = \vec{F}_{\pm} + \frac{1}{2}(\Delta_{\vec{F}_1}^{\pm} + \Delta_{\vec{F}_2}^{\pm})
\end{equation}
}
since $\vec{F}_{\pm}$ is constant, $\Delta_{\vec{F}_j}$ are both constant on half of the dataset and the two knowledge related terms have uniformly alternating sign. After centering, we have:
{\small
\begin{align*}
\widetilde{\mathcal{M}}(x_i^+) - \widetilde{\mathcal{M}}(x_i^-) = &\pm \vec{F}_{\top/\bot} \pm \Delta_{\vec{F}_j}^{\top}\\
&+ \alpha \frac{1}{2}(\Delta_{\vec{F}_1}^{\pm} - \Delta_{\vec{F}_2}^{\pm})
\end{align*}
}
where $\alpha = 1$ if $j = 1$ and $-1$ otherwise.

A probe using any of these remaining terms will have low CCS loss, with a bias towards the most salient terms. This is true for any undesirable feature that would remain in the contrast differences, and it affects both trained CCS probes and analytical CRC-TPC probes. This work aims to address this issue by removing unwanted features \textit{before} training the probe.

\section{Method}
\label{sec:method}

\begin{figure*}[htpb]
    \centering
    \vspace{-12pt}  
    \includegraphics[width=1\textwidth]{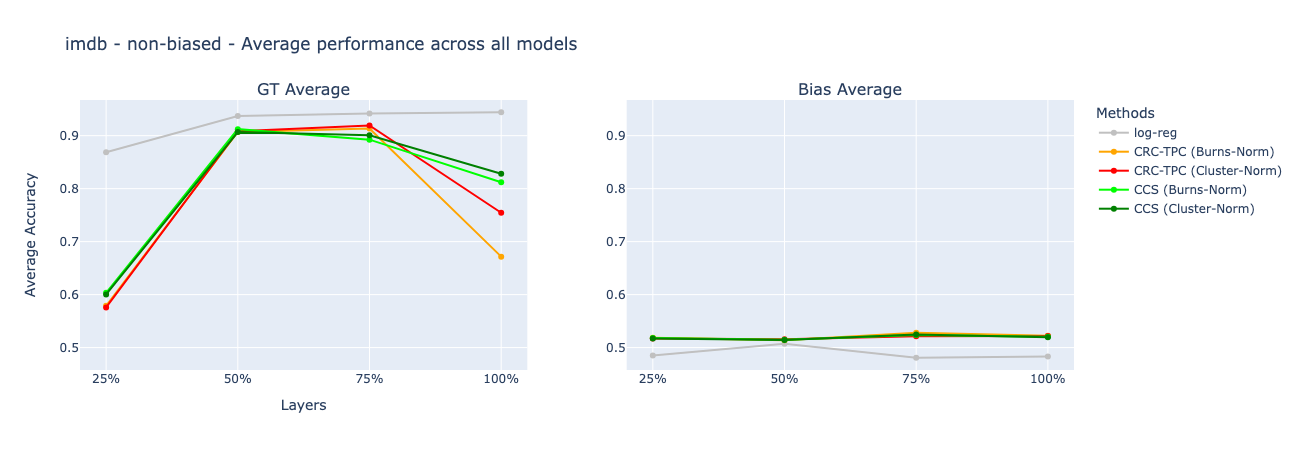}
    
    \vspace{-12pt}  
    
    \includegraphics[width=1\textwidth]{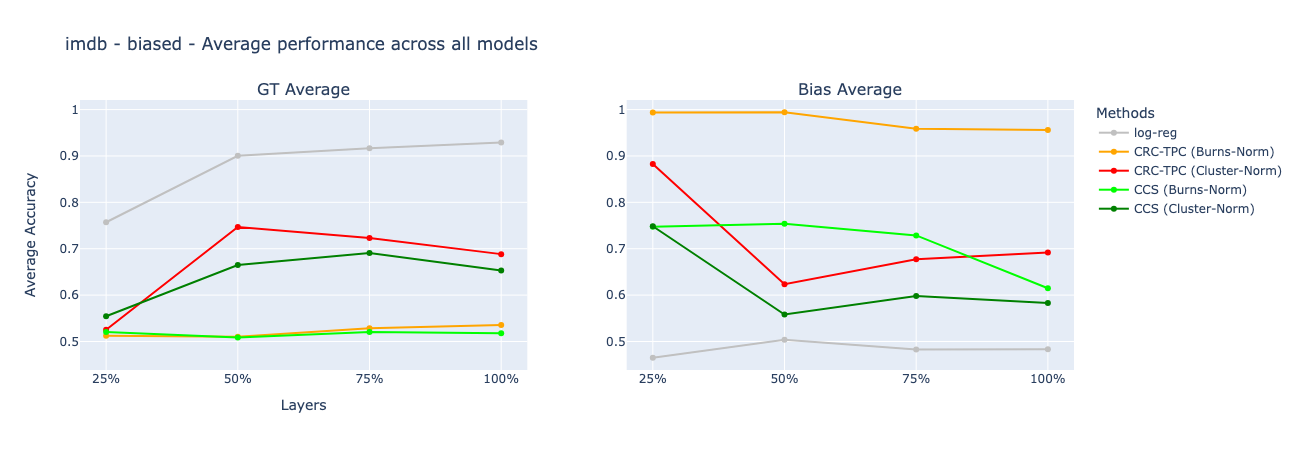}
    
    \caption{Mean accuracy of Logistic Regression, CRC-TPC, and CCS probes across six different models for (top) unmodified prompts and (bottom) modified prompts with distracting random words. Especially for the modified prompts, unsupervised methods using our Cluster-Normalization consistently outperform standard Burns-Normalization across the 25th, 50th, and 75th percentile layers and the final layer.}
    
    \label{fig:exp_1_average}
\end{figure*}

We begin with a dataset of contrast pairs, $\{(x_i^+, x_i^-)\}_{i=1}^{n}$. For each pair, we harvest the intermediate activations of a language model $\mathcal{M}$, specifically the state of the residual stream at the final token position at a specific layer, which we denote as $\mathcal{M}(x_i^{\pm})$. We average these activations for each contrast pair $\mathcal{M}(x_i) = \frac{\mathcal{M}(x_i^+) + \mathcal{M}(x_i^-)}{2}$, and partition $\{\mathcal{M}(x_i)\}_i$ using a clustering algorithm, thereby partitioning the original dataset using its most salient features. Each cluster is then normalized separately to have zero mean and unit variance, i.e. for each positive sample $x_i^+$, where $x_i$ belongs to cluster $c$, $\widetilde{\mathcal{M}}(x_i^+) = \frac{\mathcal{M}(x_i^+) - \mu_{c}^+}{\sigma_{c}^+}$, where $\mu_{c}^+$ and $\sigma_{c}^+$ are the mean and standard deviation of all positive samples in cluster $c$. The same normalization process is applied to all negative samples. Finally, an unsupervised probe can be trained on the contrast pair differences of the normalized (by cluster) samples. This approach allows the probe to isolate the desired knowledge feature, ignoring other distracting features isolated to each original cluster.

Following the notation in Section \ref{sec:theoretical_background}, if $x_i$ belongs to cluster $c \in \{1, 2\}$, a successful cluster normalization will leave:

{\small
$$\widetilde{\mathcal{M}}(x_i^+) - \widetilde{\mathcal{M}}(x_i^-) = \pm \vec{F}_{\top/\bot} \pm \Delta_{\vec{F}_c}^{\top}.$$}

This follows from equation \ref{eq:mean}, however in this case normalization is performed over $c$ only as opposed to the whole dataset.

A key element to the effectiveness of our method is that our clustering approach does not erase the effect of the desired knowledge feature. This is achieved by clustering the averages of each contrast pair, $\mathcal{M}(x_i)$. 
As a result, clustering only isolates salient non-contrastive features, and is effectively blind to $\vec{F}_{\pm}$ and $\vec{F}_{\top/\bot}$. Normalizing positive and negative samples separately per cluster aims to ensure that all contrastive features $\vec{F}'$ related to $\vec{F}_{\pm}$ are properly normalized out - including the leaks from non-contrastive features $\vec{F}$ mixing with $\vec{F}_{\pm}$, as explained in Section \ref{sec:theoretical_background}. Note, we do not normalize out similar $\vec{F}'$ resulting from the mixing of $\vec{F}$ with $\vec{F}_{\top/\bot}$. Eventually, only contrastive features related to knowledge are kept.

\section{Experiments}

In our experiments, we utilize Mistral-7B as our main language model, harvesting activations (using the libraries from \citep{wolf-etal-2020-transformers} and \citep{nanda2022transformerlens}) at the 75th percentile layer (layer 24 for Mistral-7B) since, from preliminary experiments, we find probes achieve higher accuracies using the 50th to 90th percentile layers. We also report results using different language models (Phi-2 and 3, Gemma-7B, Llama-3-8B, Pythia-6.9B) and layers in the following subsections and Appendices to verify the efficacy of our method.
Our experiments follow the same general approach as those reported in \citet{farquhar2023challenges}, as each of these original experiments set out to demonstrate the limitations of current unsupervised probing techniques. Individual results for each model can be found in Appendix \ref{app:exp1_details}.

We present results for three experiments below. For the first and second, we create prompt datasets based on the IMDb dataset \citep{jiang2023mistral, maas2011learning}, while for the third we use the CommonClaim \citep{casper2023explore} dataset. We report results on a fourth experiment utilizing the DBpedia dataset \citep{lehmann2015dbpedia} in Appendix \ref{app:experiment_3}; this experiment follows on from results reported in \citet{farquhar2023challenges}, however, we find we are unable to replicate these results (on three different models) and instead obtain high accuracies for both the original method of CCS and our approach using cluster normalization. 

Activation clustering is performed using HDBScan, implemented in the \textit{scikit-learn} library \citep{kramer2016scikit}, setting a minimum number of elements in each cluster to 5 and using the Euclidean distance metric. One advantage of HDBScan over other clustering algorithms (e.g., k-means) is that the number of clusters does not need to be specified in advance. In order to examine the variance in probe performance, we report summary statistics of 50 probe fits in each experiment, and visualize the results from all.

The following experiments generally involve a comparison between an \textit{original} prompt and a \textit{modified} one, to attempt to induce a bias in an unsupervised probe. Hereon, we refer to these original prompts as \textit{unbiased} or \textit{non-biased} and modified prompts as \textit{biased}. We also refer to normalization over an entire dataset, as \textit{Burns normalization} or \textit{Burns-Norm} (See \ref{sec:theoretical_background} for more details). We refer to our alternative approach through clustering as \textit{cluster normalization} or \textit{Cluster-Norm}. Unlike the approach in \cite{burns2022discovering}, where multiple prompt templates were used, the study in \cite{farquhar2023challenges} utilized only one prompt template per dataset. Our method follows the prompt-template setup from \cite{farquhar2023challenges}. 

Each experiment utilizes a train-test split of 70\% for training and 30\% for testing. Importantly, we evaluate our unsupervised probes on a test set where Burns-Norm is applied to the test set as it was done in  \citep{burns2022discovering}, and not our cluster normalization. This is because we want probes to generalize, so if during evaluation they are fed with a contrast pair that belongs to an entirely different dataset, it is out of distribution for the clusters found during training. The probe should be a feature in the unaltered latent space. Although we do not use cluster normalization on the test set for the aforementioned reason, we do use Burns-Norm for being able to compare our results with \citet{burns2022discovering} as this is what they do for the test set. \citet{farquhar2023challenges} likely follow a similar approach, as they mention utilizing normalization but do not provide any details regarding a train-test split.

For each experiment, we also report results using the CRC-TPC method as an alternative unsupervised probing technique to CCS. Finally, we report an upper-bound by using the results of the supervised method of logistic regression, similar as done by \cite{burns2022discovering} and later also by \citep{farquhar2023challenges}. For additional experiment details see also Appendix \ref{app:exp1_details}.

\subsection{Random Words}
\label{sec:experiment_1}

\begin{figure*}
    \centering
    \includegraphics[width=0.48\textwidth]{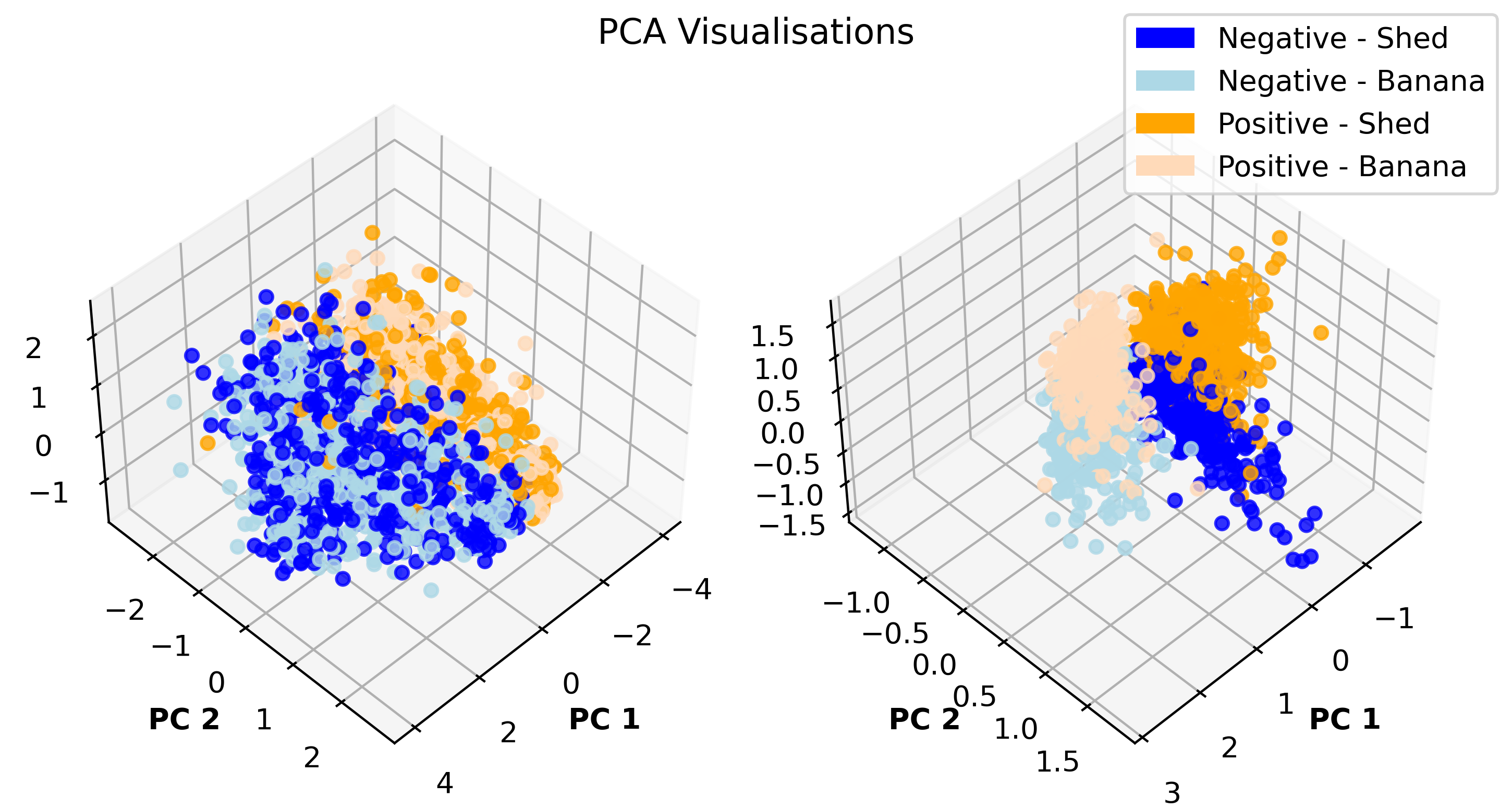}
    \hfill 
    \includegraphics[width=0.48\textwidth]{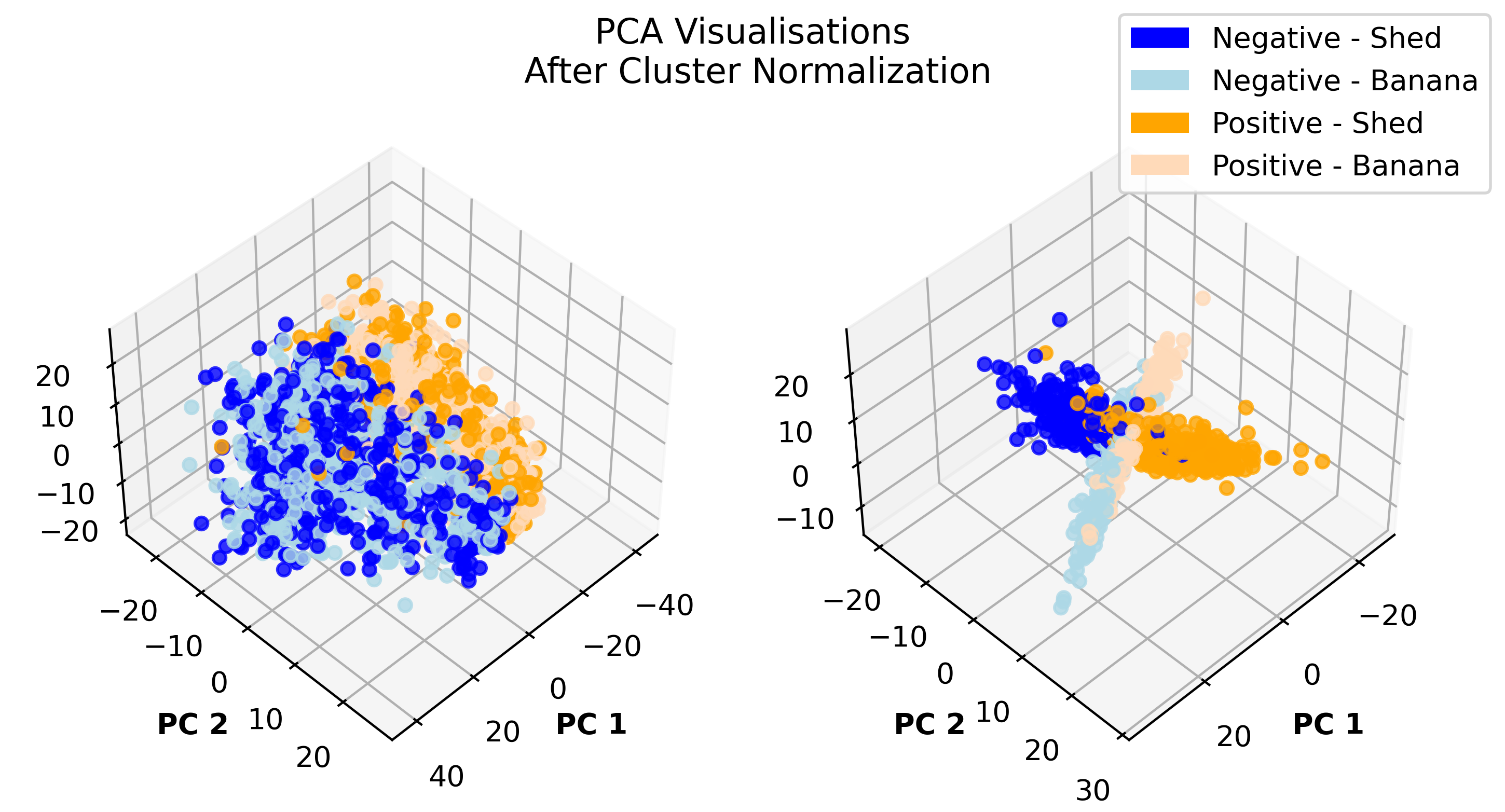}
    \caption{Visualization of the top three principal components (PCs) of the normalized contrast pair differences $\widetilde{\mathcal{M}}(x_i^+) - \widetilde{\mathcal{M}}(x_i^-)$ - with normalization performed either over the entire dataset (left) or per cluster (right) - for the random words experiment. Points are colored orange or blue based on the ground truth label (positive/negative) and shaded light or dark based on the appended random word (banana/shed). For each subfigure, we compare PCA projections using default prompts, where no random words are appended (left) against modified prompts, where random words like ``banana'' / ``shed'' are appended (right). On the left we note the first PC classifies the undesired random word feature (light vs dark). On the right, using cluster normalization, we find the first PC classifies the desired knowledge feature (orange vs blue).}
    \label{fig:pre_norm_and_post_norm}
\end{figure*}

In this experiment, we induce a strong syntactical bias in the data to illustrate the problem of distracting salient features and demonstrate the necessity of our method for removing them.

\subsubsection{Dataset} 
 Following the approach of \citet{farquhar2023challenges}, we create a dataset by appending a random word to half of our prompts and a different random word to the remaining half. The following is an example of a prompt in a given dataset, where [label] can be \textit{positive} or \textit{negative} and [random\_word] is a random word from the NLTK corpus \citep{bird2006nltk}. For each data point we have a different movie review ([review], e.g. \textit{``This is my favorite movie ...''}):

{\small
\begin{verbatim}
Consider the following example: [review],
Between positive and negative, the
sentiment of this example
is [label]. [random_word]
\end{verbatim}
}
These random words are appended with the aim of distracting an unsupervised probe. Our cluster normalization method is able to remove these distractions (See Figure \ref{fig:pre_norm_and_post_norm}).

\subsubsection{Training and Results}
We train probes on each dataset with two partitions and random words, followed by normalization over the entire dataset as described in \citet{farquhar2023challenges}. Subsequently, we train an additional set of probes for each setting using our cluster normalization method (see Section \ref{sec:method}). We find probes trained using our method achieve a much higher accuracy on average, as shown in Table \ref{table:ccs_crc_IMDb_layer_23_biased} and Figure \ref{fig:exp1_ccs}.

\begin{table}[ht]
\centering
\begin{tabular}{l|c}
\hline
\textbf{Method} & \textbf{Accuracy} \\
\hline
Logistic Regression (Upper Bound) & 0.94 \\
CRC-TPC            & 0.51 \\
\textbf{CRC-TPC w/ Cluster Norm}  & \textbf{0.81} \\
CCS             & 0.53 \\
\textbf{CCS w/ Cluster Norm}  & \textbf{0.77} \\
\hline
\end{tabular}
\caption{Accuracy results for the random words experiment on the biased IMDb dataset using Mistral-7B.}
\label{table:ccs_crc_IMDb_layer_23_biased}
\end{table}

These results show that CCS probes trained using our cluster normalization method achieve an average accuracy of $0.77$, while CRC-TPC achieves $0.81$: both relatively high. In contrast, probes following the original CCS approach without clustering tend to perform only slightly better than random guessing. This indicates that our cluster normalization method effectively identifies and eliminates the unwanted contrastive feature from random words (see Section \ref{sec:theoretical_background}). Figure \ref{fig:pre_norm_and_post_norm} visualizes the top principal components of the contrast pair differences when using two random words, clearly illustrating the saliency of this distracting feature under the original setting (left) versus avoiding this problem through cluster normalization (right). Figure \ref{fig:exp_1_average} shows the mean accuracy for different layers across multiple models, including Mistral-7B and additional models Gemma-7B, Phi-2, Phi-3, Llama-3-8B, and Pythia-6.9B-v0. The results demonstrate that our Cluster-Norm method significantly enhances the performance of unsupervised methods.

\subsection{Explicit Opinion}
In this experiment, we examine how the inclusion of an explicit opinion from a fictional character, Alice, affects the accuracy of unsupervised probes.

\label{sec:experiment_2}
\begin{figure*}
    \centering
    \includegraphics[width=.9\linewidth]{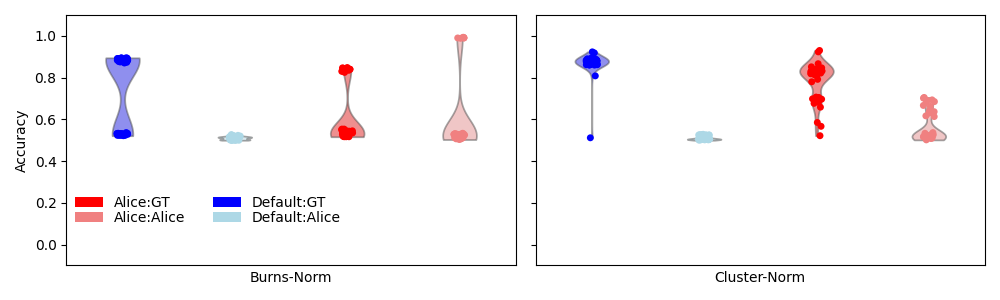}
    \caption{Discovering an explicit opinion with Mistral-7B. Accuracy when using the default prompt (blue) vs a modified prompt with the opinion of fictional Alice (red), evaluated against the ground truth sentiment labels (dark) and labels of Alice (light). Under the standard CCS approach (Burns-Norm) the case of the modified prompt, evaluated against ground truth labels (dark red) has most CCS probes achieve random accuracy. When using cluster normalization, we find this average accuracy increases.}
    \label{fig:exp2_ccs}
\end{figure*}
\subsubsection{Dataset}
Following the second experiment in \citet{farquhar2023challenges}, we create a dataset by inserting the opinion of a fictional character Alice into each prompt, with the adversarial aim of distracting an unsupervised probe into predicting the opinion of Alice. The sentiment of Alice is chosen randomly, and may or may not align with the ground truth label for a given sample - an example prompt is shown below:

{\small
\begin{verbatim}
Alice is an expert at classifying 
the sentiment of movie reviews.
Everyone agrees with her, 
and they always copy her answers.
Consider the following example:  [review]
Alice thinks it is [positive | negative].
What do you think? I think it is [label].
\end{verbatim}
}

Contrast pairs are constructed by setting [label] to either ``positive'' or ``negative''. 

\subsubsection{Training \& Results}

Our results for CCS are shown in Figure \ref{fig:exp2_ccs}. We find that a modified prompt including Alice's opinion causes the majority of our CCS probes to achieve random accuracy against ground truth labels, when normalizing over the entire dataset (Burns-Norm). Clustering before normalizing over each cluster addresses this issue - we see that the average accuracy is closer to that of the control setting, where the opinion of Alice is not inserted. (PCA visualizations analogous to those in Figure \ref{fig:pre_norm_and_post_norm} are found in Appendix \ref{app:exp_2_PCA}, while a figure displaying the mean accuracy for different
layers across multiple model can be found in  Appendix \ref{app:exp2_details}.) The reason for cluster normalization achieving higher accuracy is that our method removes the distracting feature of the opinion of Alice, enabling a CCS probe to more accurately determine the direction of the desired knowledge feature.  However, for the simpler CRC-TPC method, results differ only slightly for the two approaches, as can be seen in Table \ref{table:crc_IMDb_layer_23_biased_explicit}.

\begin{table}[ht]
\centering
\begin{tabular}{l|c}
\hline
\textbf{Method} & \textbf{Accuracy} \\
\hline
Logistic Regression (Upper Bound) & 0.85 \\
CRC-TPC            & 0.68 \\
\textbf{CRC-TPC w/ Cluster Norm}  & \textbf{0.69} \\
CCS             & 0.56 \\
\textbf{CCS w/ Cluster Norm}  & \textbf{0.77} \\

\hline
\end{tabular}
\caption{Accuracy results for the explicit opinion experiment on the biased IMDb dataset using Mistral-7B.}
\label{table:crc_IMDb_layer_23_biased_explicit}
\end{table}

Interestingly, for Mistral-7B, our results differ from those in \cite{farquhar2023challenges}. Using the default normalization method on the modified dataset, only a few CCS probes are influenced by the explicit opinion of Alice. However, results for other models we have tested (detailed in Appendix \ref{app:exp2_details}) show that the explicit opinion of Alice is indeed often a distraction for the unsupervised probes using only Burns-Normalization, though not as significantly as reported in \cite{farquhar2023challenges}.

\subsection{Prompt Template Sensitivity}
\label{sec:agent_simulation}

 \citet{farquhar2023challenges} outline two key issues with current approaches to unsupervised probing for knowledge in language models. Thus far, we have primarily discussed the first of these issues: distracting salient features can satisfy the CCS loss, and trained probes may align with these unwanted features instead of the desired knowledge feature. Our approach of cluster normalization addresses this limitation, as demonstrated in the above experiments.

The second issue - frequently discussed in the eliciting latent knowledge literature \citep{christianoELK2021} - has to do with the quality of the desired knowledge feature itself, and how to discriminate between a model's true knowledge and that of a simulated agent. Prompting can induce a language model into a state of simulating a different quality of knowledge. For example, performance on difficult question-answering datasets can be improved by prompting language models to \textit{``provide Professor Smith's answers to a question''} \citep{evans2021truthful}: it is implied that Professor Smith is an intelligent agent highly capable of answering such questions. In \citet{farquhar2023challenges}, it is observed that current techniques for unsupervised probing are unable to detect a model's capability to answer a given question under such a prompt change. 

In the interest of clearly defining the scope of our cluster normalization method, we carry out an additional experiment investigating prompt template sensitivity, and demonstrate that our method also does \textbf{not} give rise to uniform scores across different prompts. Therefore, our method does not address this second issue.

\subsubsection{Datasets}

We use the CommonClaim \citep{casper2023explore} dataset, which assesses models' abilities to identify true or false statements, with many questions deliberately designed to trick the model through common misconceptions and traps. 

In this experiment, the language model is presented with a question and possible answer, and asked to evaluate the truth of the answer. In order to evaluate prompt template sensitivity, we consider three settings:
\begin{itemize}
    \item A \textit{default} setting, where the model is simply asked to evaluate the truth of a possible answer to a given question.
    \item A \textit{professor} setting, where the following text is prepended to each prompt, before asking the question: ``\textit{Professor Smith was given the following instructions: Interpret each claim literally, and as a question about the real world; carefully research each claim, without falling prey to any common myths; and reply `I have no comment' unless you are completely certain of the answer. Here are the Professor’s responses:}''
    \item A \textit{literal} setting, where the first and last phrase of the above text are removed, to investigate whether the invocation of a specific character is important.
\end{itemize}

An example prompt in this set-up is shown below:
{\small
\begin{verbatim}
Professor Smith was given...
Question: [question]
Answer: [answer]
Is this answer true or false?
\end{verbatim}
}
Contrast pairs are constructed by appending either ``True'' or ``False'' to each prompt.

\subsubsection{Training and Results}

Each question is presented with a possible answer (our dataset contains an equal random split of true and false answers) under each of the three settings: \textit{default}, \textit{literal}, and \textit{professor}. Three sets of unsupervised probes are then trained, one for each setting, and are scored on their ability to identify a feature corresponding to ground truth labels. We compare performance of normalizing over the entire dataset, as in \citet{burns2022discovering}, to our cluster normalization approach.

\begin{figure*}[htbp]
    \centering
    \includegraphics[width=.9\linewidth]{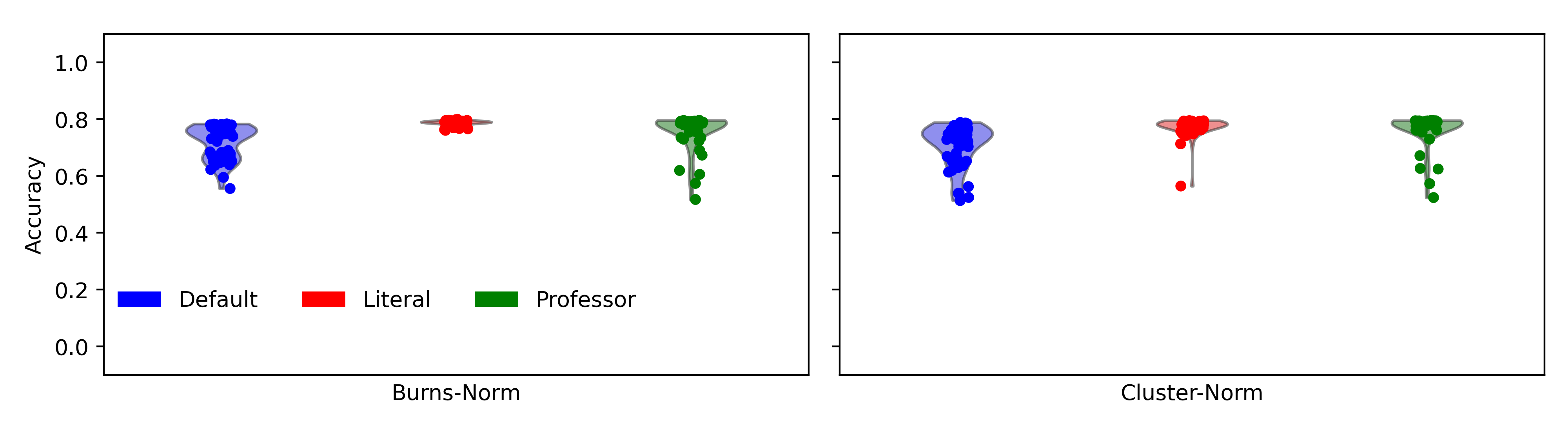}
    \caption{Variation in probe accuracy when investigating prompt template sensitivity using the CommonClaim dataset, for Mistral-7B. In the default setting (blue), when compared to the literal (red) and professor (green) settings, we see a slightly more varied spread in probe accuracy, regardless of the use of cluster normalization.}
    \label{fig:exp4_ccs}
\end{figure*}

CCS probe accuracies are visualized in Figure \ref{fig:exp4_ccs}. We see that in the default (blue) setting, the variance in probe accuracy is slightly higher than the literal (red) or professor (green) settings. Indeed, this difference is also clear when we examine the performance of CRC-TPC, shown with the average performance of all probing methods in Table \ref{table:exp4_crc}.

\begin{table}[ht]
\centering
\resizebox{\columnwidth}{!}{
\begin{tabular}{l|rrr}
\hline
 & \textbf{Default} & \textbf{Literal} & \textbf{Professor} \\
\hline
Logistic Regression (Upper Bound) & 0.81  & 0.81  & 0.81 \\
CRC-TPC & 0.66 & 0.79 & 0.79 \\
CRC-TPC w/ Cluster-Norm & 0.66 & 0.79 & 0.79 \\
CCS & 0.66 & 0.73 & 0.76 \\
CCS w/ Cluster-Norm & 0.65 & 0.74 & 0.76 \\
\hline
\end{tabular}
}
\caption{Average accuracy of different probing techniques when investigating prompt template sensitivity using the CommonClaim dataset, for Mistral-7B. For all unsupervised probing methods we see a lower accuracy in the default prompt setting when compared to the other two, regardless of the use of cluster normalization. Logistic regression is included as an upper-bound.}
\label{table:exp4_crc}
\end{table}

Notably, these findings remain regardless of the use of cluster normalization, for both CCS and CRC-TPC. Cluster normalization offers no concrete benefit here, as there are no distracting features to be removed. Rather, the knowledge feature itself exhibits different qualities due to the prompt.

This experiment illustrates when cluster normalization is and is not helpful. Cluster normalization offers a solution to the issue of distracting features, but does not yield a method of unsupervised probing which is robust to prompt changes i.e., differentiation between general knowledge and simulated knowledge. Experimental results using an alternative dataset (TruthfulQA \citep{evans2021truthful}) and different models, can be found in Appendix \ref{app:exp3_details}.

\section{Related Work} 

It has been shown that language models develop internal representations of the world \citep{li2022emergent}, with individual concepts often encoded as linear directions in activation space \citep{elhage2022toy,nanda2023emergent, burns2022discovering, marks2023geometry}. Language models can also output false information, even if the encoded \textit{knowledge} in the activations seems to indicate a correct internal representation of the information \citep{evans2021truthful, azaria2023internal, campbell2023localizing}. We seek to elicit this latent knowledge \citep{christiano2022eliciting} in an unsupervised manner. In recent years several methods have been proposed \citep{burns2022discovering, belrose2023eliciting, belroseVinc2024, zou2023representation, li2024inference}, although unsupervised methods can be subject to undesirable biases, as shown by \citet{farquhar2023challenges}. They demonstrate that unsupervised probing techniques, such as those developed in \citet{burns2022discovering}, often identify the most salient features in a dataset, as opposed to knowledge only. These features may not always align with the specific knowledge feature of interest, as described in Section \ref{sec:method}. We provide theoretical explanations for some of these issues, and propose a method to eliminate them.

The work we cite in the introduction and background sections focuses on finding a general linear representation of knowledge in the latent space of a language model. While we focus on unsupervised approaches, most work concentrates on supervised ones \citep{christiano2022eliciting, marks2023geometry}. This body of work is part of a more general field of research that aims at ensuring truthfulness of language models, by making sure that what they answer is actually what they believe or follows from reasoning e.g., working with quirky language models or using chain-of-thought reasoning \citep{turpin2023language, lyu2023faithful, radhakrishnan2023question, mallen2023eliciting}.

\section{Discussion and Conclusion}
In this study, we address significant challenges associated with the unsupervised probing of knowledge in language models. The primary issue tackled is that of distracting salient features that can mislead the probing process. Our cluster normalization technique shows promising results in effectively isolating and minimizing the impact of such distractions, thereby enhancing the performance of unsupervised probes. Our results demonstrate that without proper normalization, probes tend to align with the most salient features present in the dataset, which are not necessarily related to the target knowledge feature. This observation mostly aligns with findings from previous studies \citep{farquhar2023challenges}, which showed that unsupervised probes are prone to capturing irrelevant features when such features are salient. However, in general, our results do not show as pronounced an effect as \citep{farquhar2023challenges} suggested for the standard CCS method. This observation is especially true for the experiments detailed in Section \ref{sec:experiment_2} and Appendix \ref{app:experiment_3}). Nonetheless, through cluster normalization, we provide a promising method to mitigate the issue of distracting salient features by identifying these features and ensuring that they are canceled out during the training of the probe. This normalization allows the probe to focus more accurately on the intended knowledge feature.

\section{Limitations}

Our study also highlights limitations of current probing techniques that are not addressed by our method. Specifically, as noted by \citet{farquhar2023challenges}, we find that prompting techniques which can induce a language model into simulating a different \textit{quality} of knowledge by simulating an agent can still affect our unsupervised probe performance. This is a critical limitation, as we specifically want to elicit the knowledge of the model, not that of some simulated entity. Addressing this limitation is another significant challenge for the research community, as it requires an investigation into the question of whether a language model's knowledge as its capacity to answer a given question under \textit{any} prompt differs from simulated knowledge, and whether such a difference could be exploited to increase the reliability of probing algorithms. These limitations are studied in \citet{mallen2023eliciting}, where the context-dependence of knowledge probes is measured.

Another potential limitation of our method is that, as mentioned in Section \ref{sec:method}, it relies on the fact that the mean of each pair of activations contains no information related to knowledge, which seems to be the case in practice but may need to be further investigated.

Further research is also needed to explore the effect of the choice of basis on probing algorithms, using e.g. the Local Interaction Basis developed by \citet{bushnaq2024using} or overcomplete bases given by dictionary learning \citep{cunningham2023sparse, braun2024identifying}.

\section{Acknowledgements}
We would like to thank Alex Mallen, Erik Jenner, Clément Dumas, Paul Colognese, Steffen Thoma, Tilman Räuker, Joseph Bloom, and Achim Rettinger for their valuable feedback and discussions. We also extend our gratitude to the Long Term Future Fund and Manifund for supporting Kaarel and Walter during part of this work, and to the Graduate School of Computer Science at Université Paris-Saclay for supporting Grégoire. Additionally, Walter received support from the FZI Research Center for Information Technology for part of this work, for which he is grateful. Sharan is supported by the UKRI Centre for Doctoral Training in Application of Artificial Intelligence to the study of Environmental Risks [EP/S022961/1].

\bibliography{main}

\appendix
\onecolumn


\newpage
\paragraph{}

\section{Implicit Opinion}
\label{app:experiment_3}
\begin{figure*}[htbp]
    \centering
    \includegraphics[width=.9\linewidth]{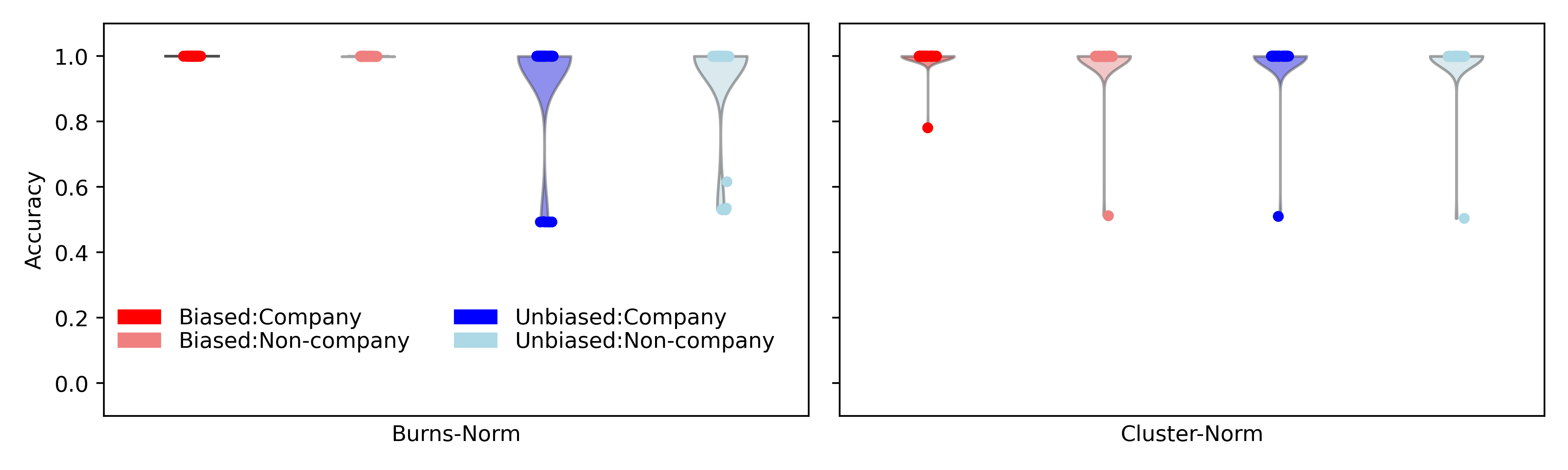}
    \caption{Discovering an implicit opinion for Mistral-7B. Biased (red) and unbiased (blue) for company (dark) and non-company (light) data. In contrast to the findings in \citet{farquhar2023challenges}, we find that most CCS probes in all settings perform with high accuracy.}
    \label{fig:exp3_ccs}
\end{figure*}

In this section, we discuss an additional experiment from \citet{farquhar2023challenges} that we are unable to replicate, achieving comparably high outcomes using both the original CCS method and our approach with Cluster-Norm.

In practical future use-cases of unsupervised probing, it is unlikely that distracting salient features will be present as explicitly as in Sections \ref{sec:experiment_1} and \ref{sec:experiment_2}. In this experiment, the character of Alice is included and her opinion (acting as the distracting feature) is implied rather than explicitly stated. In \citet{farquhar2023challenges} it is demonstrated that this distracting feature is frequently selected by CCS probes. We investigate the use of cluster normalization to address this issue. 

\subsection{Dataset}

We use the DBpedia topic classification dataset \citep{lehmann2015dbpedia} to construct a task: the language model must identify the correct topic of a given text sample, given two answer choices. An example prompt from this task is shown below:
\begin{verbatim}
    Alice answers questions about the 
    topic of text passages.
    Example: [text sample]
    Choice 1: [choice 1]
    Choice 2: [choice 2]
    Alice says "Choice [choice]
\end{verbatim}

Contrast pairs are constructed by changing the value of [choice] to 1 or 2. 

In order to convey Alice's implied opinion, few shot examples are included in the prompt, in a biased and unbiased (as a control) manner. In the latter case, the correct choice is always given in few shot examples. In the former, Alice is shown to always choose incorrectly when the correct label is ``company''. A specific few shot example used in this biased setting is shown below:
\begin{verbatim}
    Example: Thai Beverage is a large 
    drinks company in Asia.
    Choice 1: album
    Choice 2: company
    Alice says "Choice 1". Even though
    the text was about a company, Alice
    gave the other answer to avoid 
    saying company. Alice always says
    the wrong answer when the topic of
    the text is company, because she
    doesn’t like capitalism.
\end{verbatim}

Under this experimental setting, should Alice's biased implicit opinion act as a distracting feature for a CCS probe, we would notice a drop in probe accuracy for the correct answer in the biased setting, specifically on questions with the correct answer ``company''. For further details on this experiment, including the exact few-shot prompts used, see \citet{farquhar2023challenges}.

\subsection{Results}

\begin{figure*}[htbp]
    \centering
    \includegraphics[width=0.7\linewidth]{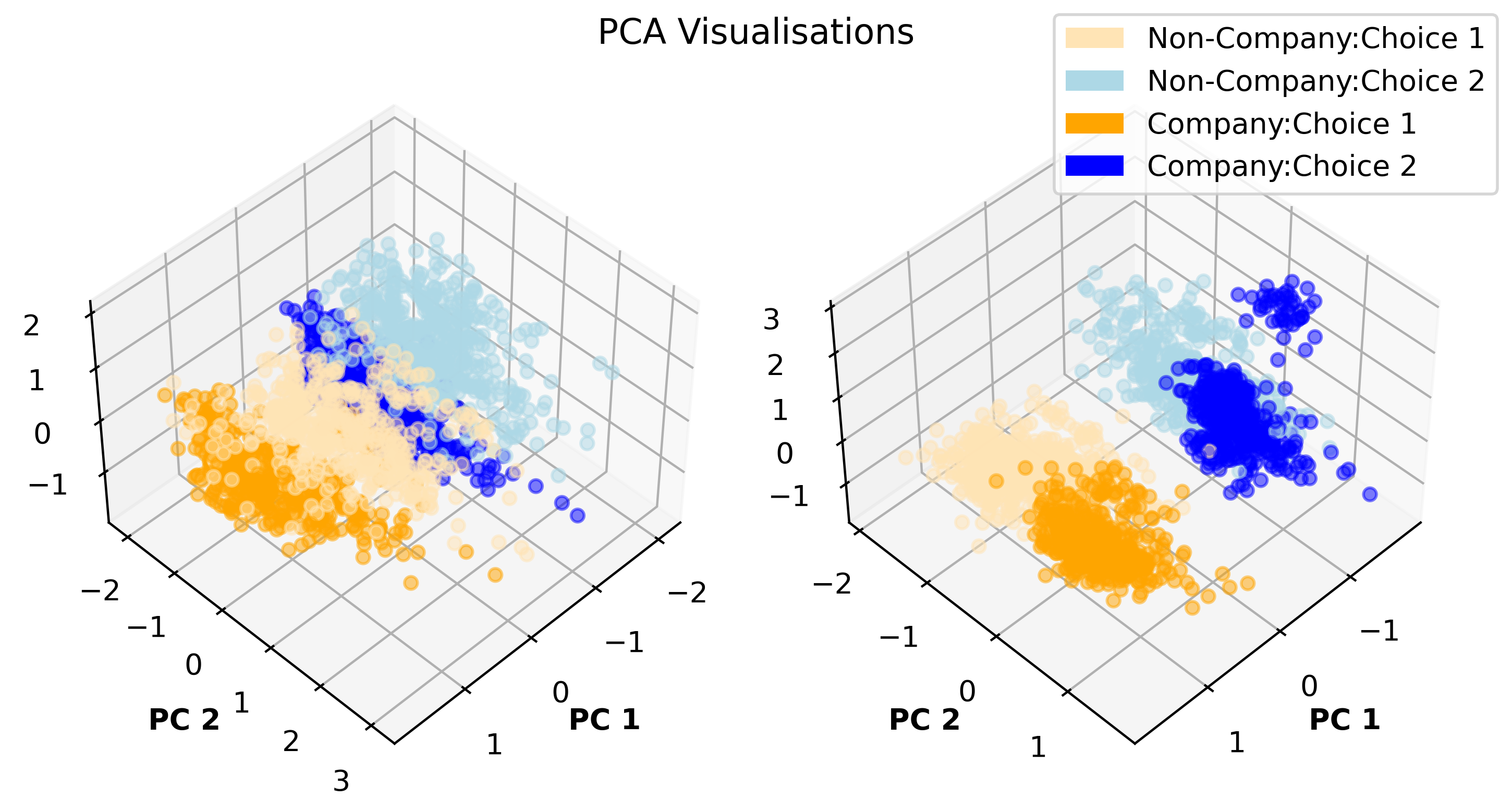}
    \caption{PCA visualizations for the implicit opinion experiment for Mistral-7B. In both the unbiased (left) \textit{and} biased (right) settings, we find that the first principal component can split the data relatively easily into two clusters, representing when the correct choice is 1 (orange) and when the correct choice is 2 (blue).}
    \label{fig:exp3_pca}
\end{figure*}

For CCS, we examine probe accuracy in four different settings: biased and unbiased, each on either questions where the correct answer was ``company'' and questions where the correct answer was not ``company''.

Our results, shown in Figure \ref{fig:exp3_ccs}, differ from those in \citet{farquhar2023challenges} in a few ways. We find that generally speaking, CCS probes in all settings of this experiment perform with high accuracy, notably including the biased setting on ``company'' data, even when using the original CCS method (Burns normalization). A small number of probes achieve roughly random accuracy, but importantly, we find that no probes in the biased setting on ``company'' data achieve (close to) zero accuracy. In other words, the feature of Alice's implied anti-company opinion is never selected by our CCS probes.

The technical reasoning for this is clarified when visualizing the harvested activations, projected onto their first three principal components, as shown in Figure \ref{fig:exp3_pca}. We see, in both the unbiased and biased cases, that the first principal component's projection can classify activations into those where the correct choice is 1 (orange) and those where the correct choice is 2 (blue) with relative ease. This is reflected in the performance of CRC-TPC on these data, shown in table \ref{table:exp3_crc}.

\begin{table}[ht]
\centering
\begin{tabular}{l|rr}
\hline
\textbf{Setting} & \textbf{Company} & \textbf{Non-company} \\
\hline
Biased & 1.00 & 1.00 \\
Unbiased & 1.00 & 0.96 \\
\hline
\end{tabular}
\caption{\textbf{CRC-TPC} performance for the implicit opinion experiment. In Figure \ref{fig:exp3_pca} we see the first principal component splits correct answers relatively cleanly, so high accuracy here is unsurprising.}
\label{table:exp3_crc}
\end{table}

The question still remains as to the reason for the differing results here, when compared to those in \citet{farquhar2023challenges}. We believe the most likely reason is model size: while we report results using Mistral-7B, \citet{farquhar2023challenges} make use of Chinchilla 70B: a much larger model. The PCA visualizations in Figure \ref{fig:exp3_pca} show that at our model size, the feature of Alice's biased opinion is not salient i.e., it is not represented by the model as cleanly as the ``correct choice'' feature, and it is for this reason that our CCS probes never select the implicit opinion feature. Regardless, this results in an inability to compare the original CCS method with cluster normalization.

\section{Additional Probing Results}
\label{app:exp_details}
In addition to Mistral-7B, the random word experiment and explicit opinion experiment is repeated for the following models: Gemma-7B, Phi-2 and 3, Llama-3-8B and Pythia-6.9B-v0. We harvested activations at four different points: the 25th percentile layer, 50th, 75th and the last layer. Unbiased examples correspond to probes trained on the original prompts, while biased examples correspond to probes trained on the modified prompts.

\subsection{Random Words}
\label{app:exp1_details}

\begin{figure*}[htbp]
    \centering
    \includegraphics[width=1\textwidth]{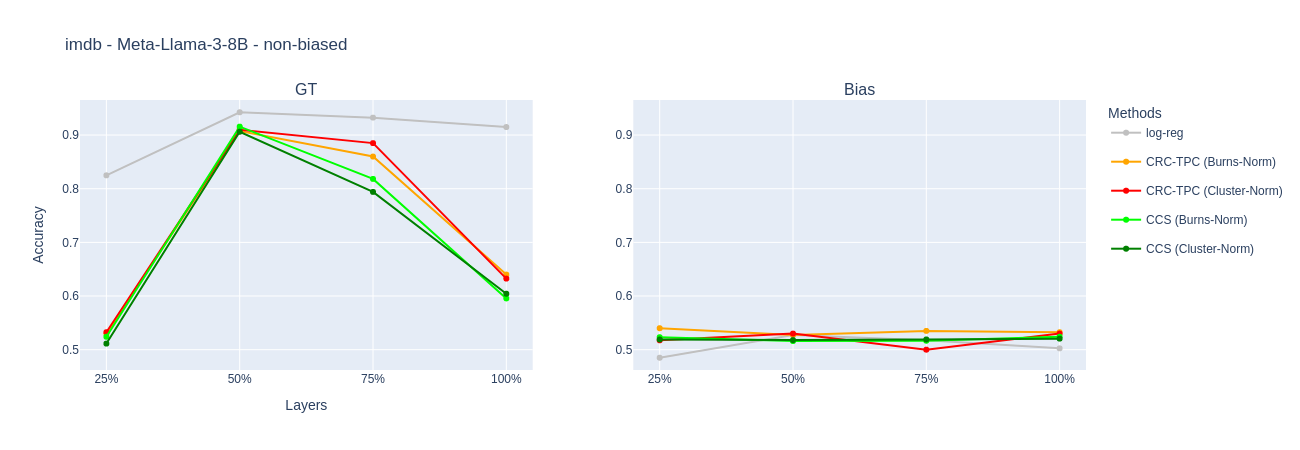}
    \includegraphics[width=1\textwidth]{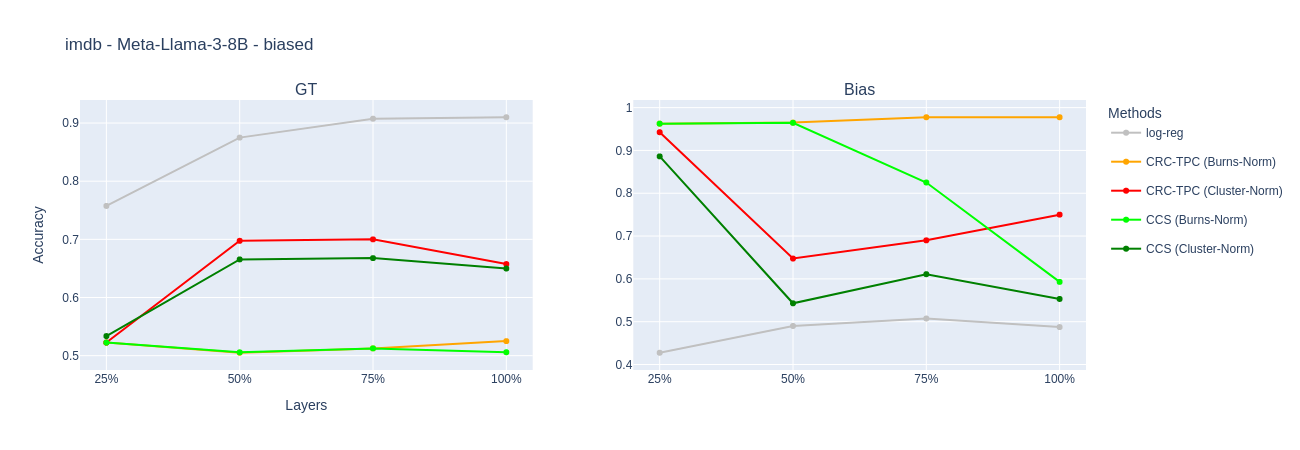}
    \caption{Mean accuracy of Logistic Regression, CRC and CCS probes on Llama-3-8B on original prompts (up) and biased ones (down) for the random word experiment.}
    \label{fig:exp_1_llama}
\end{figure*}

\begin{figure*}[htbp]
    \centering
    \includegraphics[width=1\textwidth]{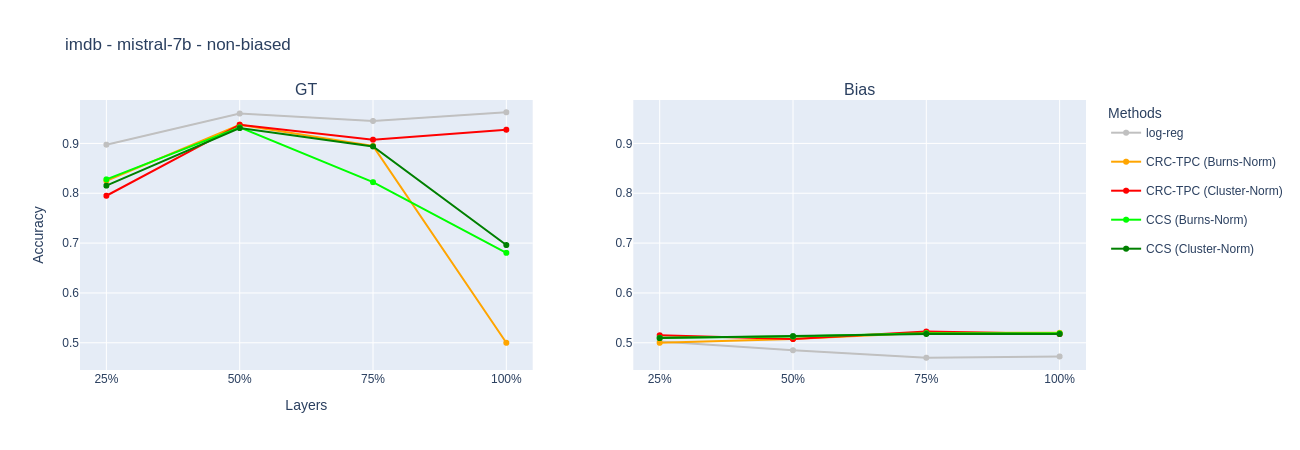}
    \includegraphics[width=1\textwidth]{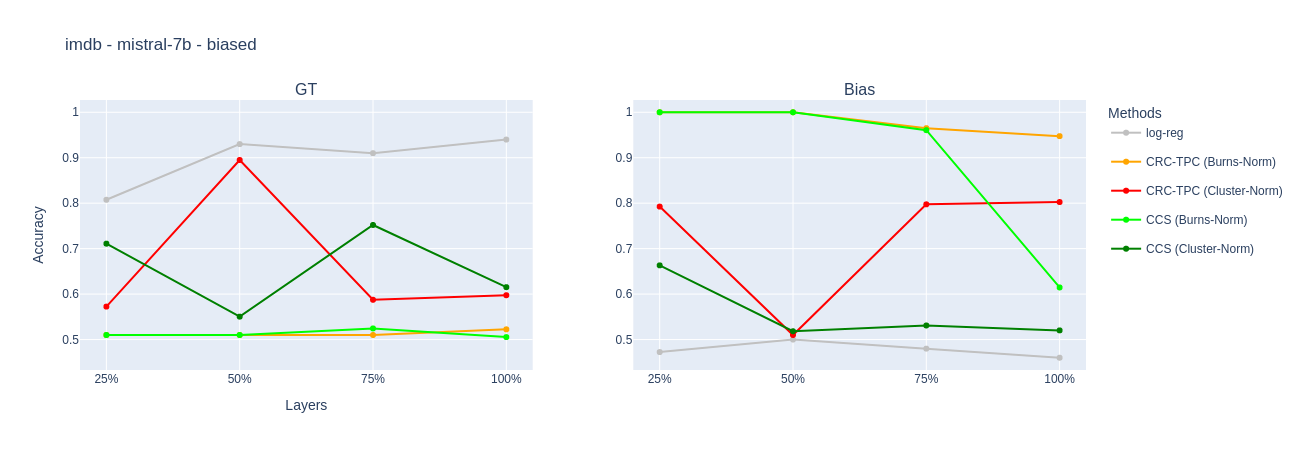}
    \caption{Mean accuracy of Logistic Regression, CRC and CCS probes on Mistral-7B on original prompts (up) and biased ones (down) for the random word experiment.}
    \label{fig:exp_1_mist}
\end{figure*}

\begin{figure*}[htbp]
    \centering
    \includegraphics[width=1\textwidth]{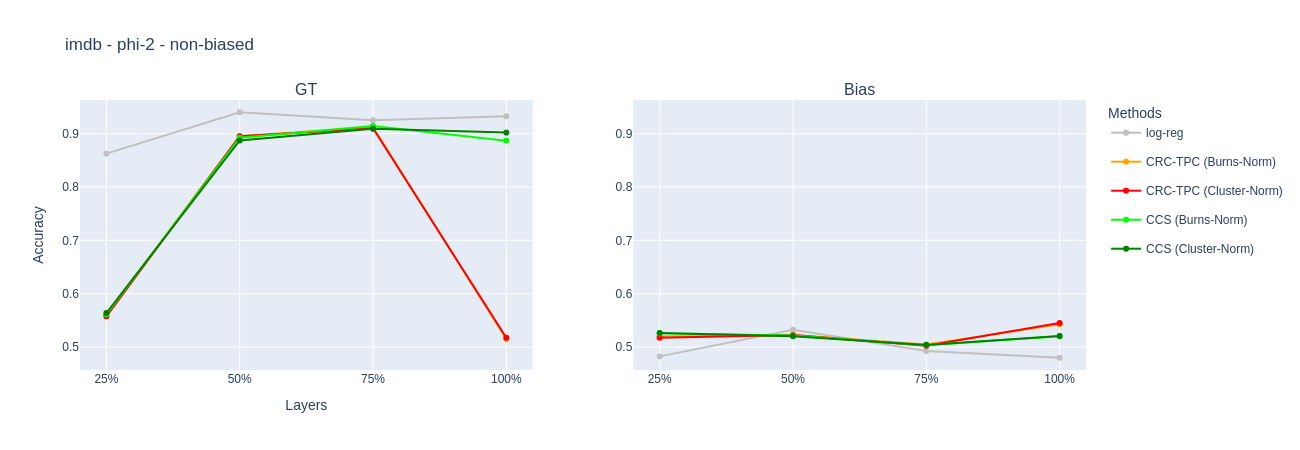}
    \includegraphics[width=1\textwidth]{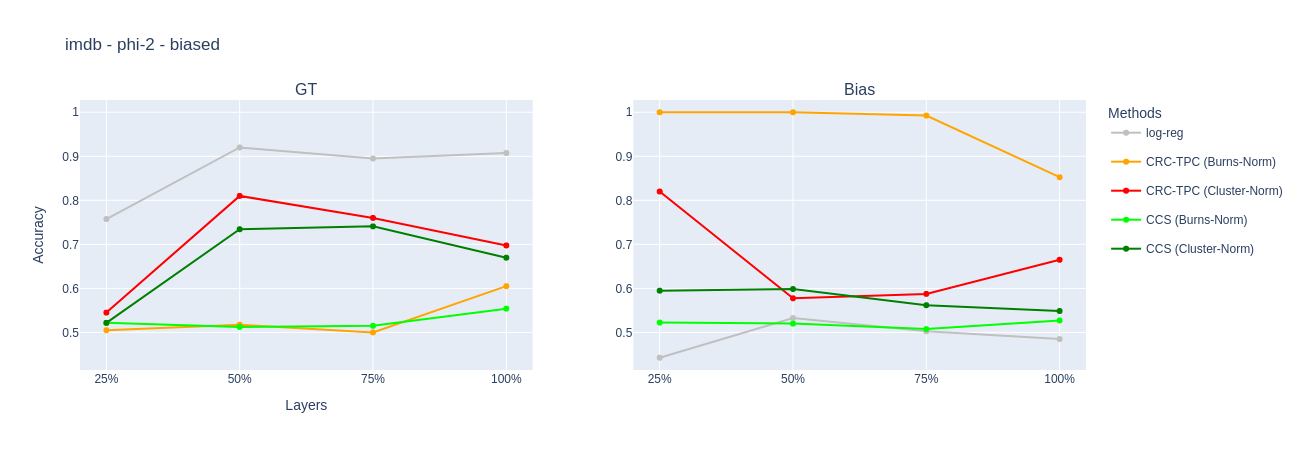}
    \caption{Mean accuracy of Logistic Regression, CRC and CCS probes on Phi-2 on original prompts (up) and biased ones (down) for the random word experiment.}
    \label{fig:exp_1_phi}
\end{figure*}

\begin{figure*}[htbp]
    \centering
    \includegraphics[width=1\textwidth]{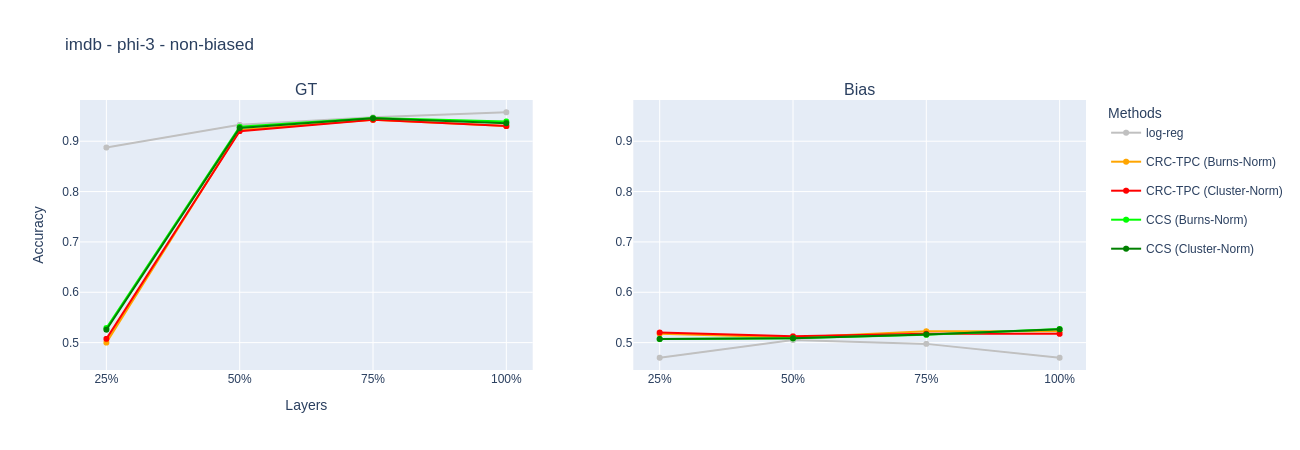}
    \includegraphics[width=1\textwidth]{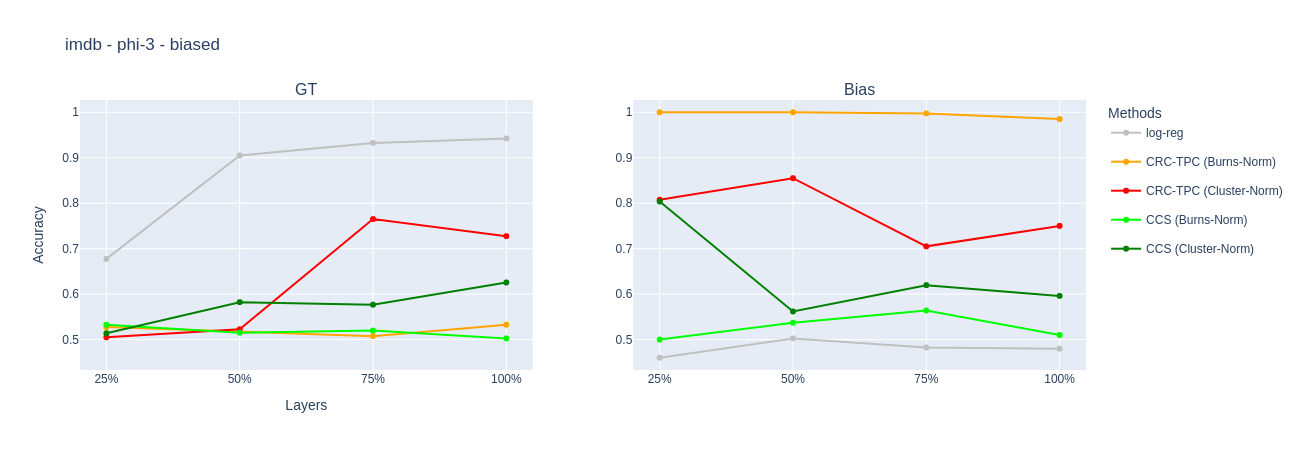}
    \caption{Mean accuracy of Logistic Regression, CRC and CCS probes on Phi-3 on original prompts (up) and biased ones (down) for the random word experiment.}
    \label{fig:exp_1_phi_3}
\end{figure*}

\begin{figure*}[htbp]
    \centering
    \includegraphics[width=1\textwidth]{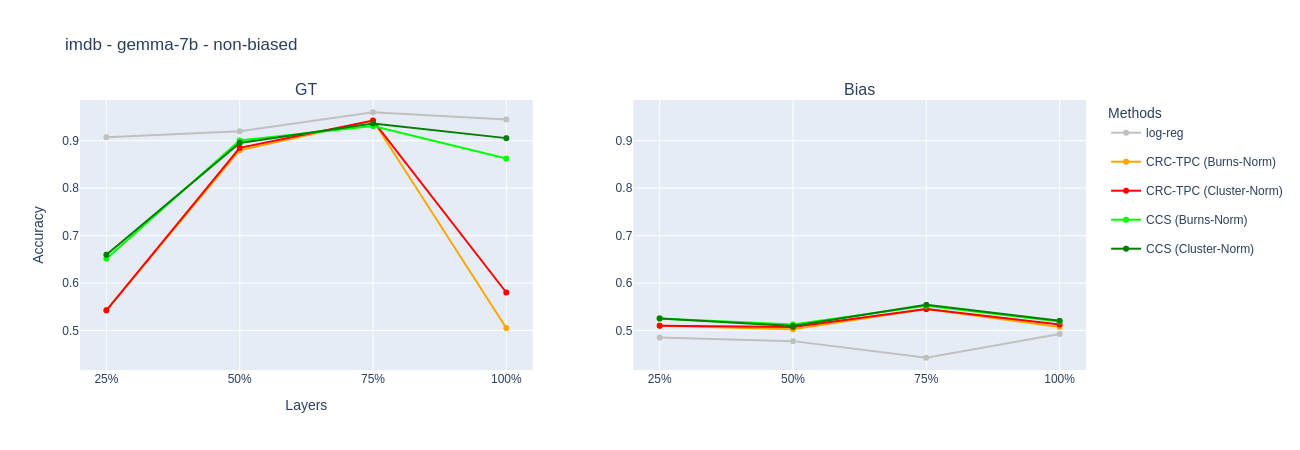}
    \includegraphics[width=1\textwidth]{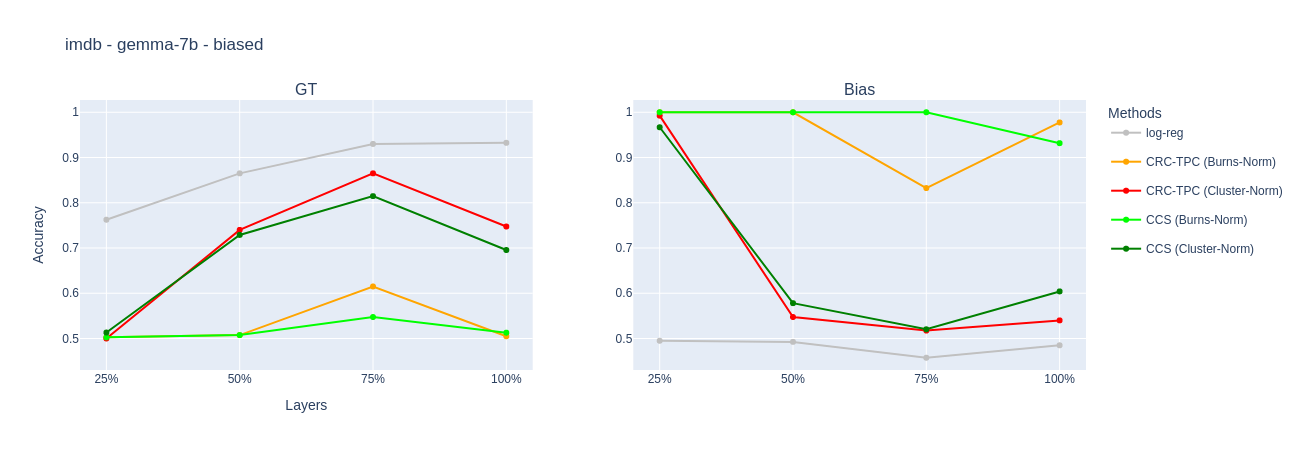}
    \caption{Mean accuracy of Logistic Regression, CRC and CCS probes on Gemma-7B original prompts (up) and biased ones (down) for the random word experiment.}
    \label{fig:exp1_gemma}
\end{figure*}

\begin{figure*}[htbp]
    \centering
    \includegraphics[width=1\textwidth]{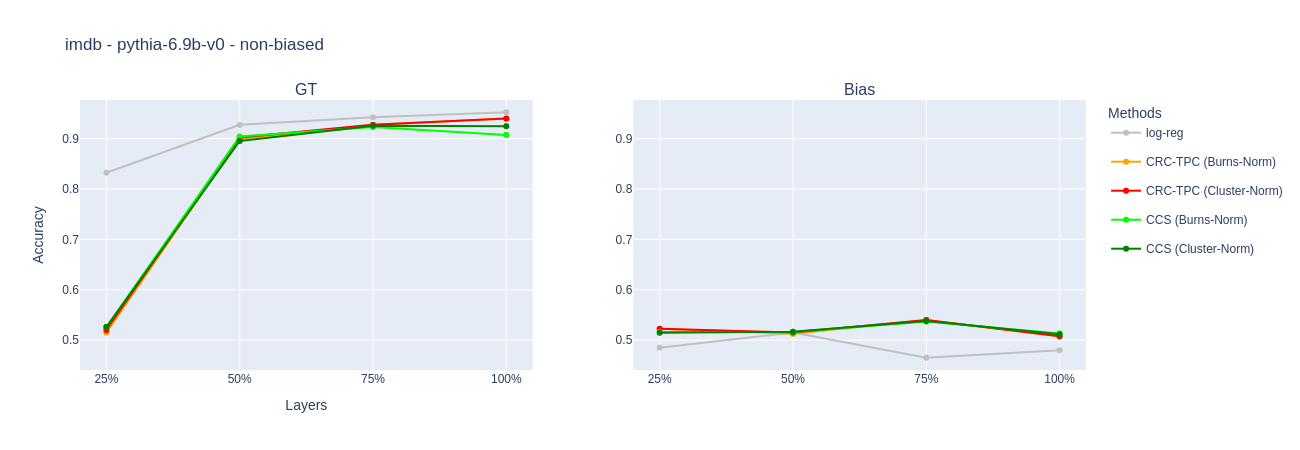}
    \includegraphics[width=1\textwidth]{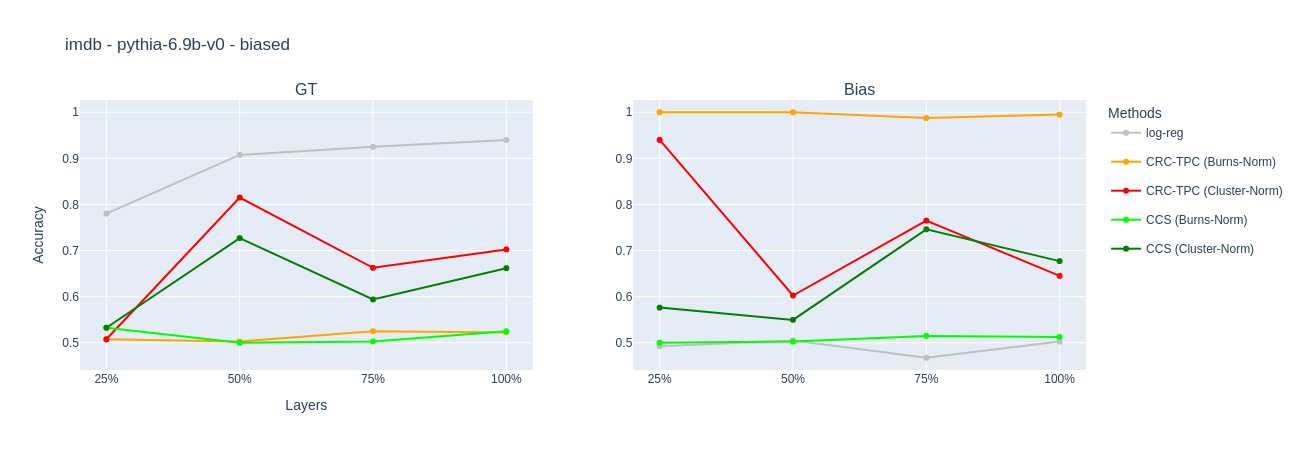}
    \caption{Mean accuracy of Logistic Regression, CRC and CCS probes on Pythia-6.9B-v0 original prompts (up) and biased ones (down) for the random word experiment.}
    \label{fig:exp_1_pythia}
\end{figure*}

The results for the additional models and layers are comparable to those of Mistral-7B at the 75th percentile layer. For the biased prompt-template dataset, unsupervised methods using cluster normalization do usually perform better than those using the standard Burns-Normalization.  
Figures \ref{fig:exp_1_llama}, \ref{fig:exp_1_mist}, \ref{fig:exp_1_phi}, \ref{fig:exp_1_phi_3}, \ref{fig:exp1_gemma}, and \ref{fig:exp_1_pythia} show violin plots of the results for the additional models.

\subsection{Explicit Opinion}
\label{app:exp2_details}

\begin{figure*}[htbp]
    \centering
    \includegraphics[width=0.7\linewidth]{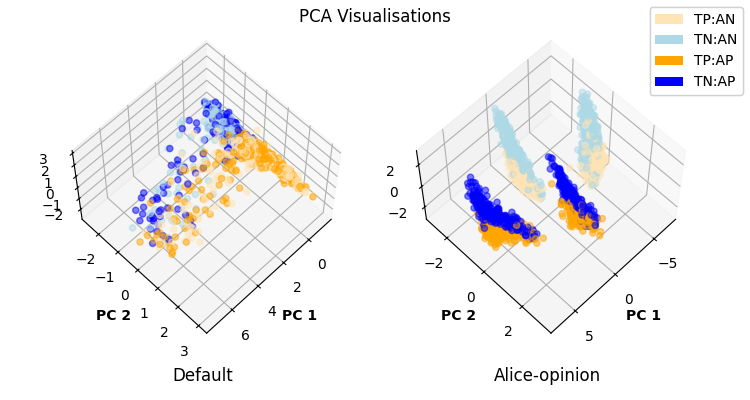}
    \caption{Visualization of the top three PC of $\widetilde{\mathcal{M}}(x_i^+) - \widetilde{\mathcal{M}}(x_i^-)$ - without per cluster normalization. Left : activations from the default prompts. Right: activations from prompts biased with Alice's opinion. TP/N : true positive/negative label, AP/N : Alice's positive or negative opinion.}
    \label{fig:exp2_pca}
    \label{app:exp_2_PCA}
\end{figure*}

Figure \ref{fig:exp2_pca} shows PCA visualizations of contrast differences without our cluster normalization, analogous to Figure \ref{fig:pre_norm_and_post_norm}. 

\begin{figure*}[htbp]
    \centering
    \includegraphics[width=1\textwidth]{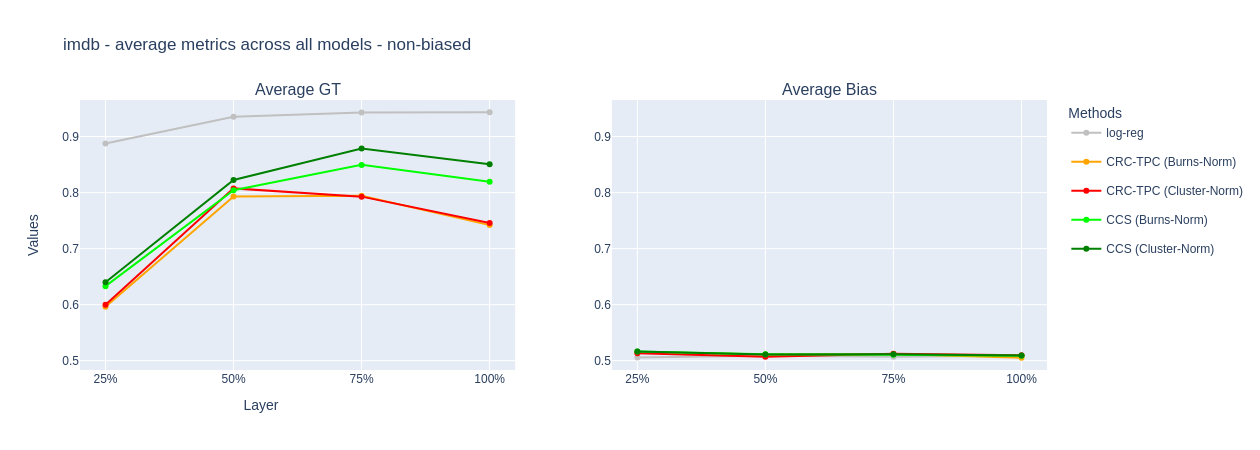}
    \includegraphics[width=1\textwidth]{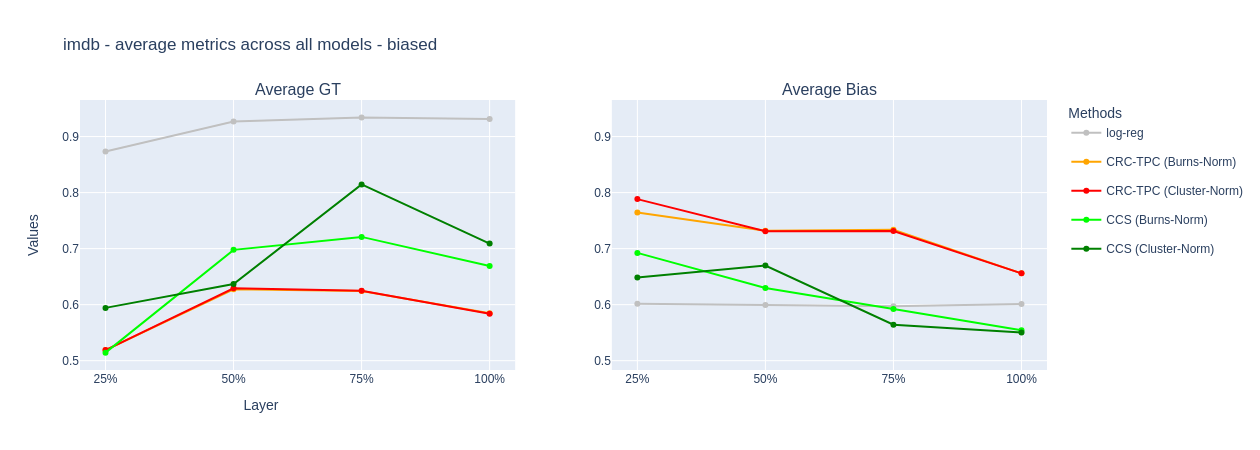}
    \caption{Mean accuracy of Logistic Regression, CRC-TPC, and CCS probes across six models for (top) original and (bottom) biased prompts in the explicit opinion experiment. CCS probes using Cluster-Normalization consistently outperform those using Burns-Normalization, particularly for modified prompts, across the 25th and 75th percentile layers and the final layer.}
    \label{fig:exp_2_average_layers}
\end{figure*}

\begin{figure*}[htbp]
    \centering
    \includegraphics[width=1\textwidth]{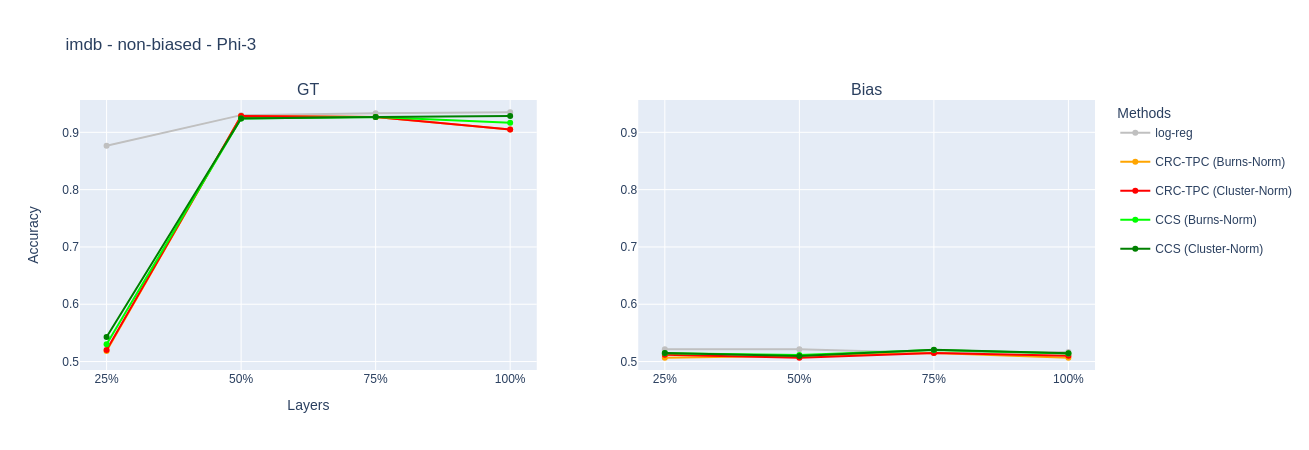}
    \includegraphics[width=1\textwidth]{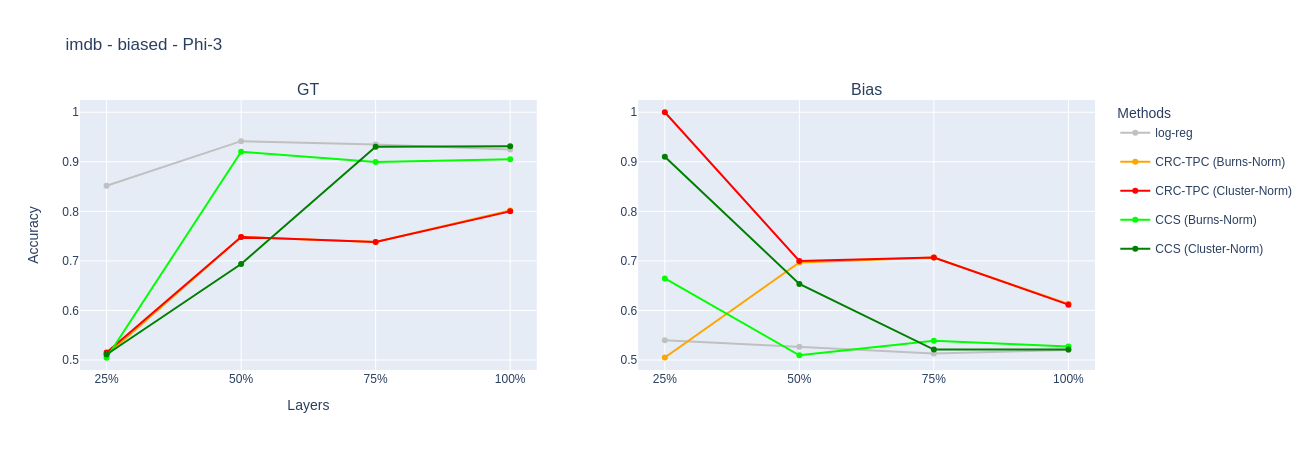}
    \caption{Mean accuracy of Logistic Regression, CRC and CCS probes on Phi-3-Instruct Mini on original prompts (up) and biased ones (down) for the explicit opinion experiment.}
    \label{fig:exp_2_phi_3_layers}
\end{figure*}

\begin{figure*}[htbp]
    \centering
    \includegraphics[width=1\textwidth]{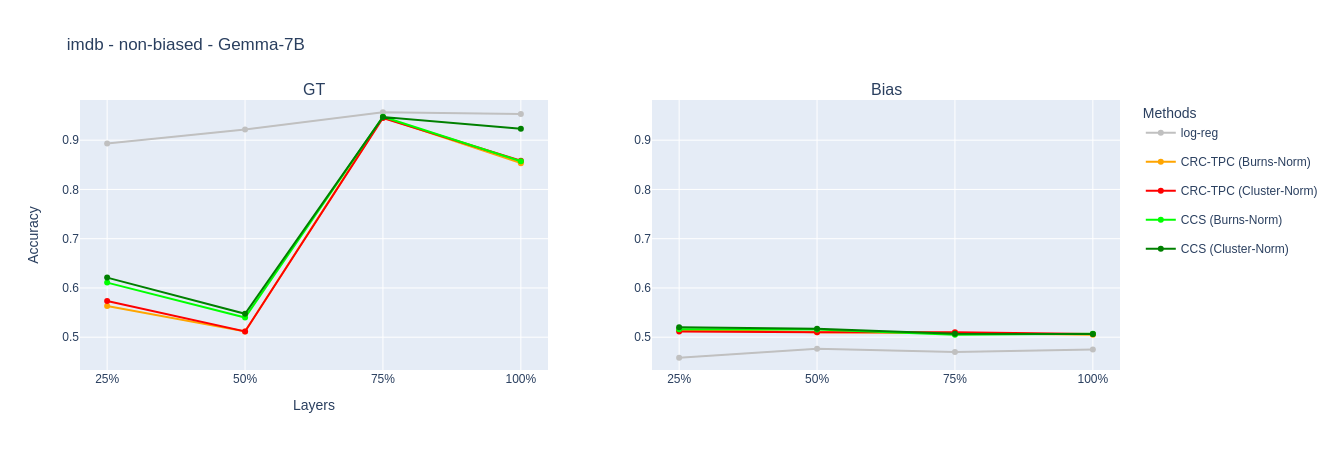}
    \includegraphics[width=1\textwidth]{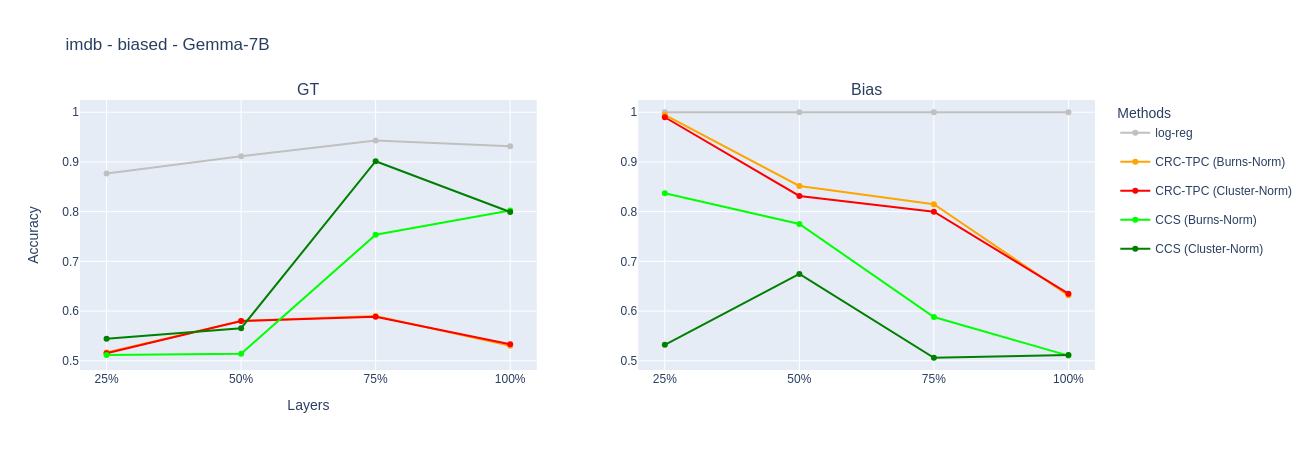}
    \caption{Mean accuracy of Logistic Regression, CRC and CCS probes on Gemma-7B on original prompts (up) and biased ones (down) for the explicit opinion experiment.}
    \label{fig:exp_2_gemma_layers}
\end{figure*}

\begin{figure*}[htbp]
    \centering
    \includegraphics[width=1\textwidth]{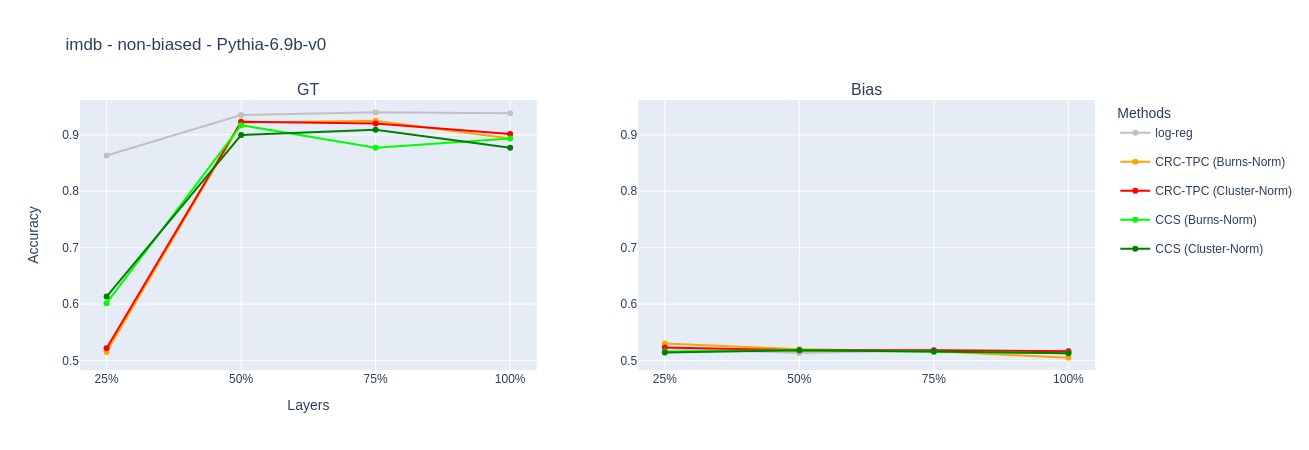}
    \includegraphics[width=1\textwidth]{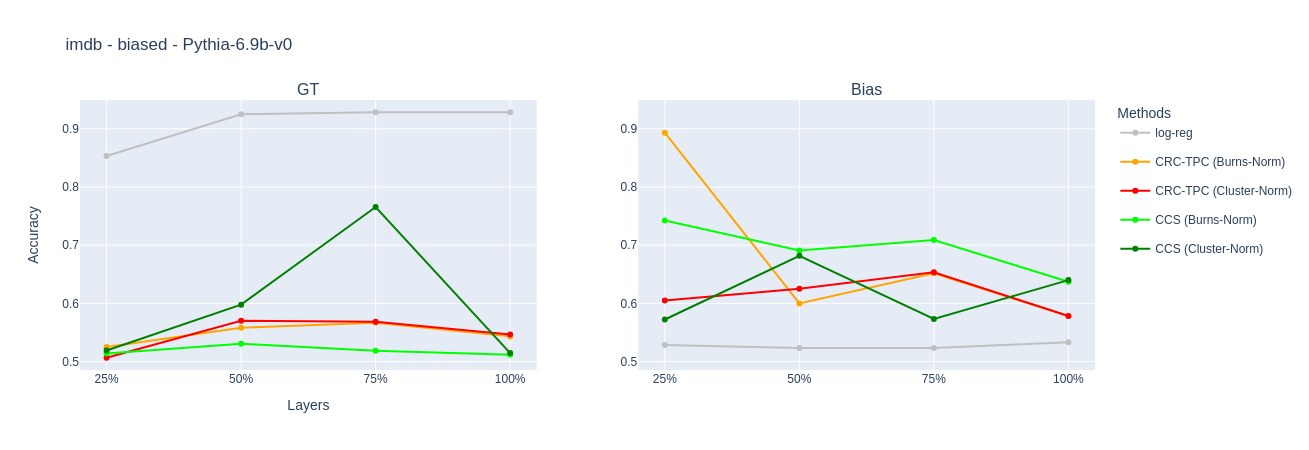}
    \caption{Mean accuracy of Logistic Regression, CRC and CCS probes on Pythia-6.9B-v0 on original prompts (up) and biased ones (down) for the explicit opinion experiment.}
    \label{fig:exp_2_pythia_layers}
\end{figure*}

The results for the additional models and layers are comparable to those of Mistral-7B at the 75th percentile layer. Figure \ref{fig:exp_2_average_layers} shows the average results across all models, including Gemma-7B, Phi-2, Phi-3, Llama-3-8B, Pythia-6.9B-v0, and Mistral-7B. For both the unbiased and biased prompt-template datasets, unsupervised methods using cluster normalization tend to outperform those using standard Burns-Normalization, with the difference being more pronounced for the biased dataset. However, the performance gap between these two methods is smaller compared to the random word experiment. Moreover, in our work, the standard CCS using Burns-Normalization appears to perform better on the biased dataset than reported by \citet{farquhar2023challenges} for the various models and layers. The individual results for the different layers and models are shown in the following figures:
\ref{fig:exp_2_mistral_layers},  \ref{fig:exp_2_phi_2_layers}, \ref{fig:exp_2_phi_3_layers}, \ref{fig:exp_2_gemma_layers} and \ref{fig:exp_2_pythia_layers}.

\begin{figure*}[htbp]
    \centering
    \includegraphics[width=1\textwidth]{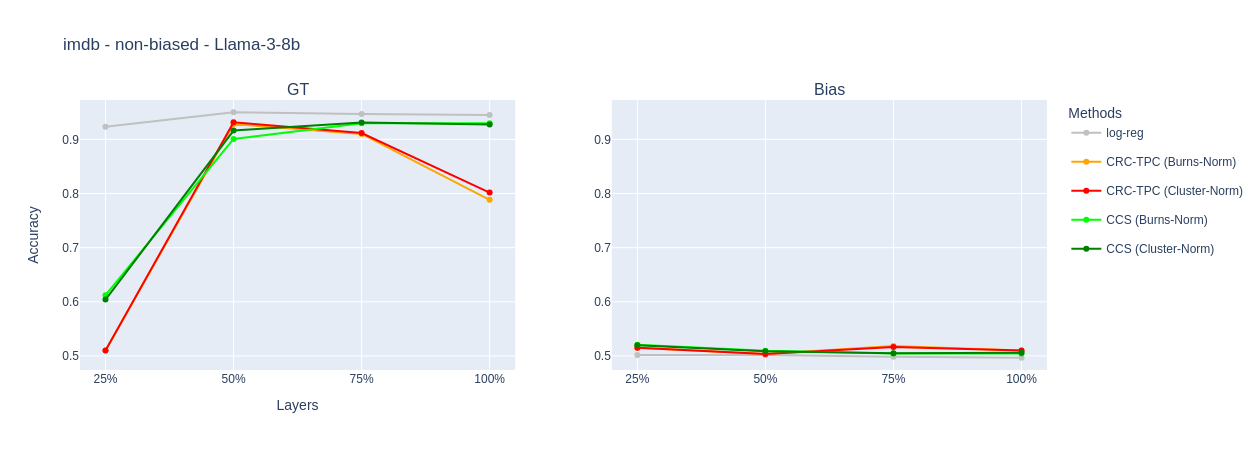}
    \includegraphics[width=1\textwidth]{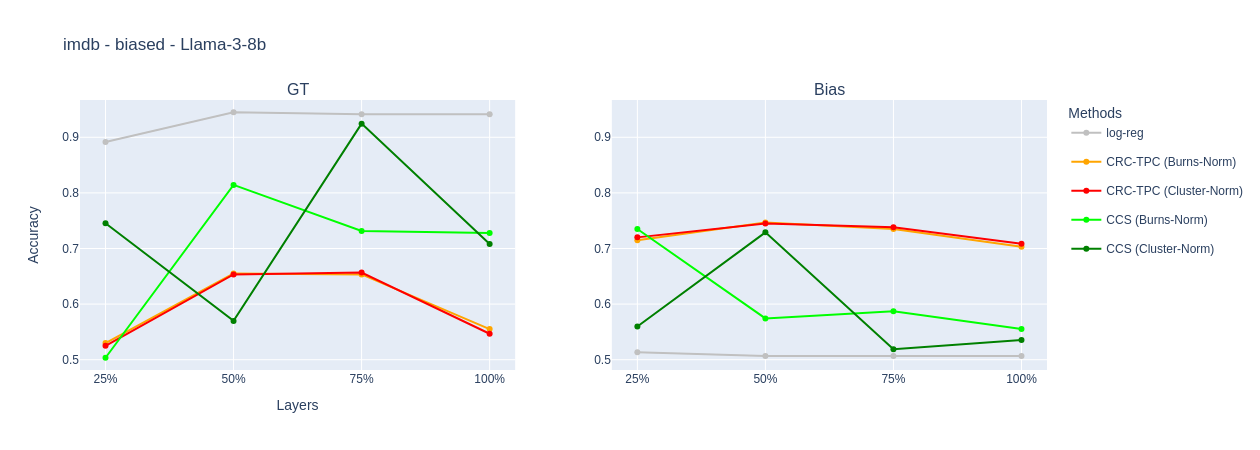}
    \caption{Mean accuracy of Logistic Regression, CRC and CCS probes on LLama-3-8b on original prompts (up) and biased ones (down) for the explicit opinion experiment.}
    \label{fig:exp_2_mistral_layers}
\end{figure*}

\begin{figure*}[htbp]
    \centering
    \includegraphics[width=1\textwidth]{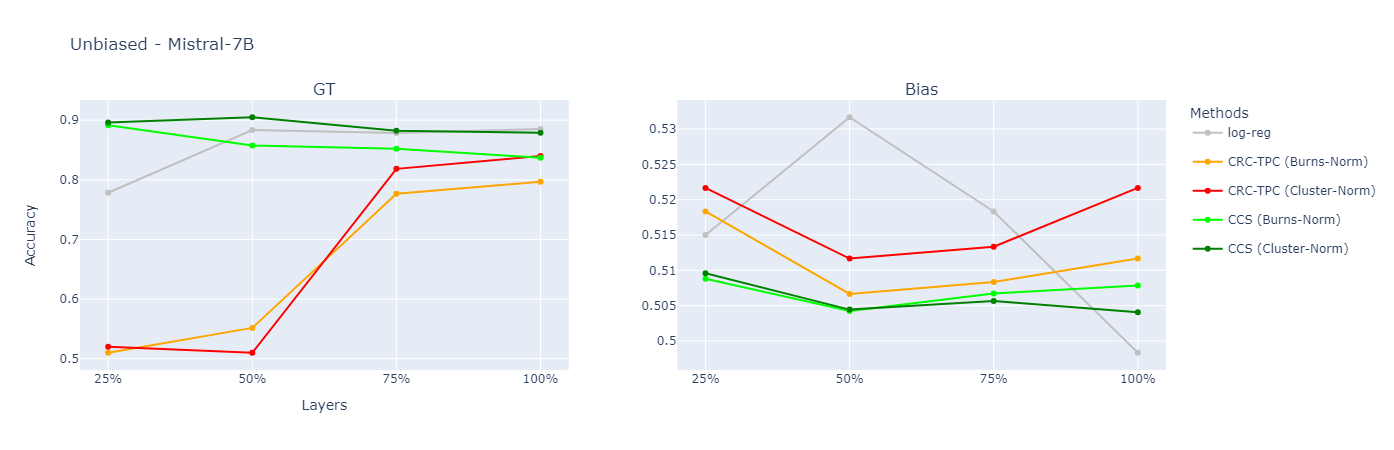}
    \includegraphics[width=1\textwidth]{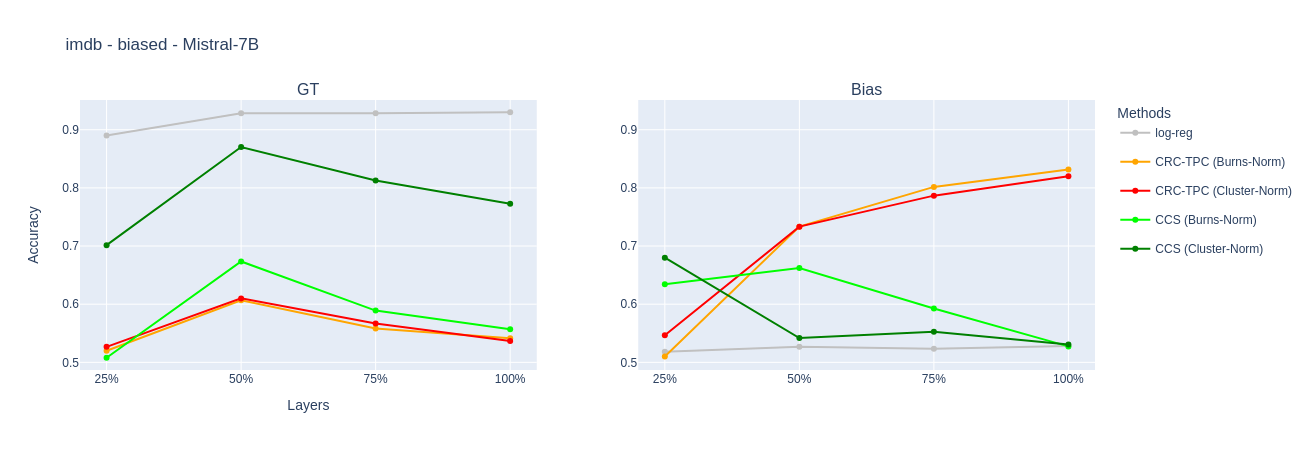}
    \caption{Mean accuracy of Logistic Regression, CRC and CCS probes on Mistral-7B on original prompts (up) and biased ones (down) for the explicit opinion experiment.}
    \label{fig:exp_2_mistral_layers}
\end{figure*}

\begin{figure*}[htbp]
    \centering
    \includegraphics[width=1\textwidth]{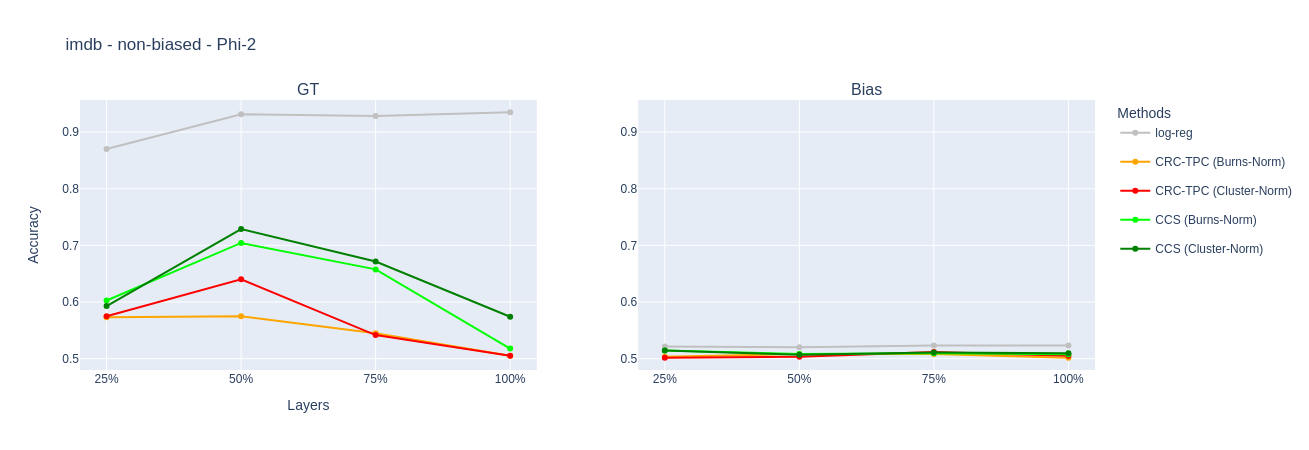}
    \includegraphics[width=1\textwidth]{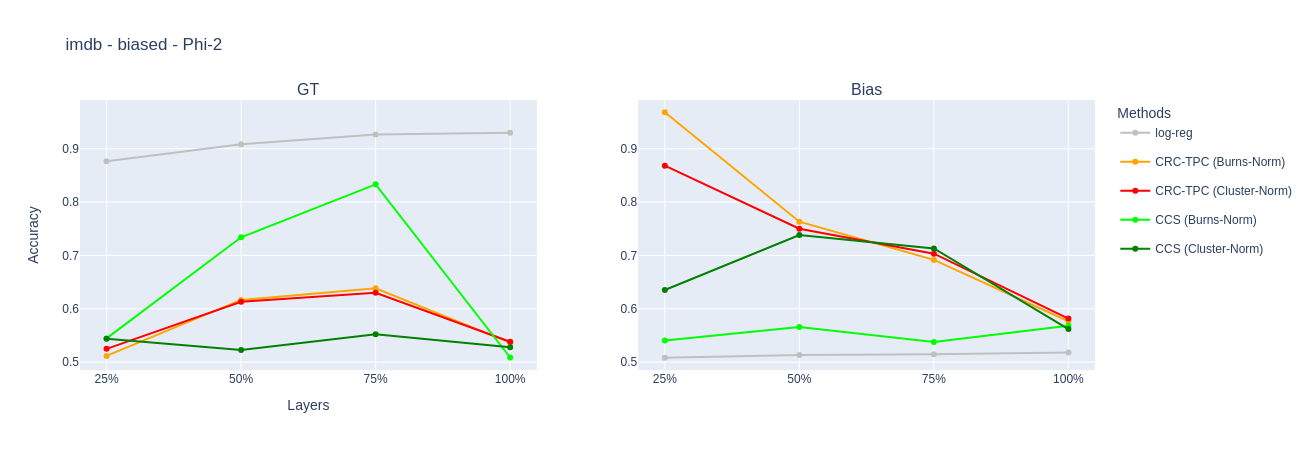}
    \caption{Mean accuracy of Logistic Regression, CRC and CCS probes on Phi-2 on original prompts (up) and biased ones (down) for the explicit opinion experiment.}
    \label{fig:exp_2_phi_2_layers}
\end{figure*}

\begin{figure*}[htbp]
    \centering
    \includegraphics[width=1\textwidth]{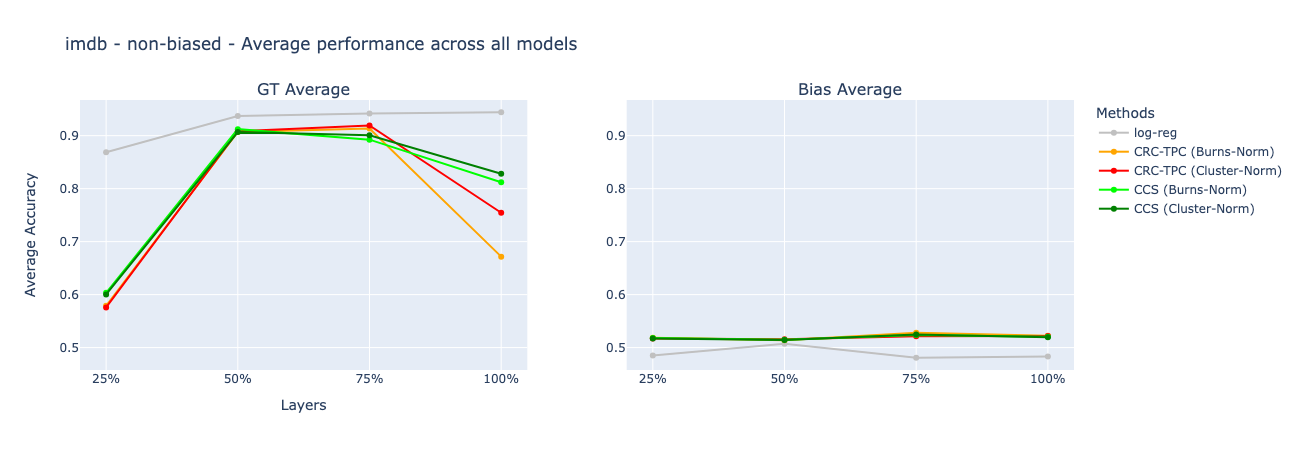}
    \includegraphics[width=1\textwidth]{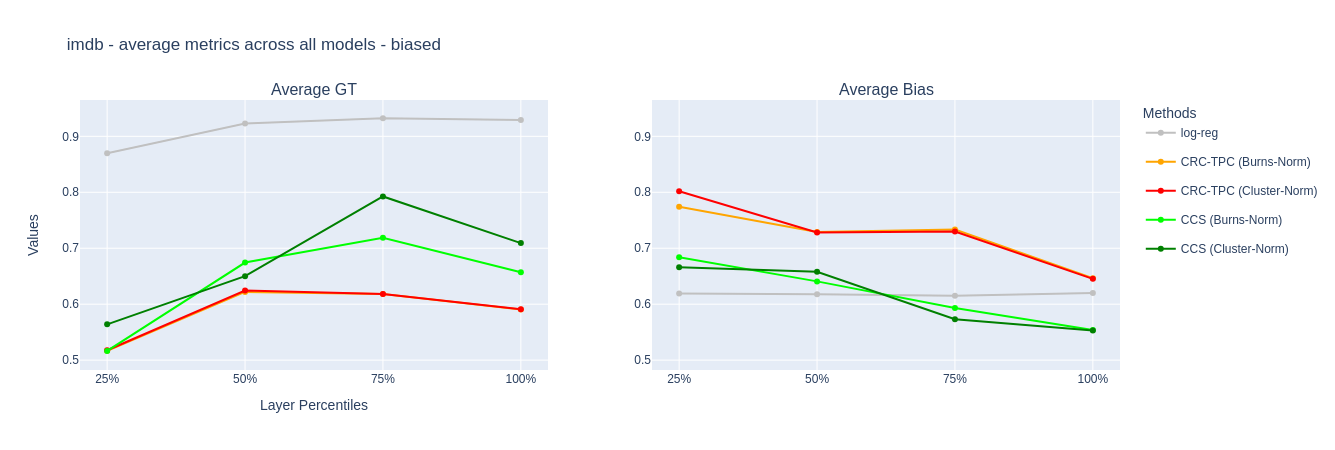}
    \caption{Mean accuracy of Logistic Regression, CRC and CCS probes averaged across all models on original prompts (up) and biased ones (down) for the explicit opinion experiment.}
    \label{fig:exp_2_average}
\end{figure*}

\subsubsection{Violin Plots — 75th Percentile Layer}

\begin{figure*}[htbp]
    \centering
    \includegraphics[width=\linewidth]{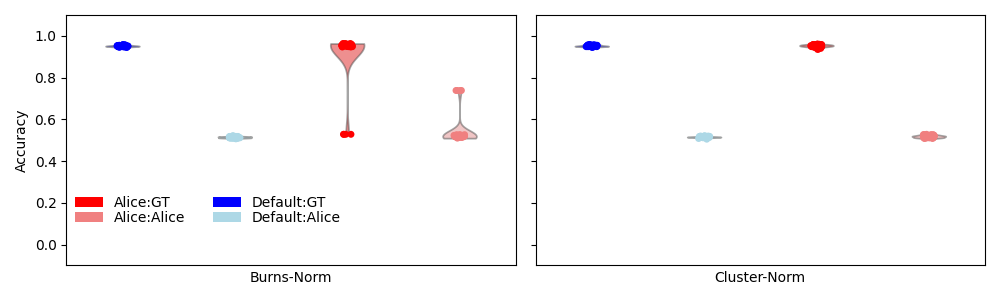}
    \caption{Phi-3 Mini - Explicit Opinion Experiment}
    \label{fig:experiment_2_phi_3}
\end{figure*}

\begin{figure*}[htbp]
    \centering
    \includegraphics[width=\linewidth]{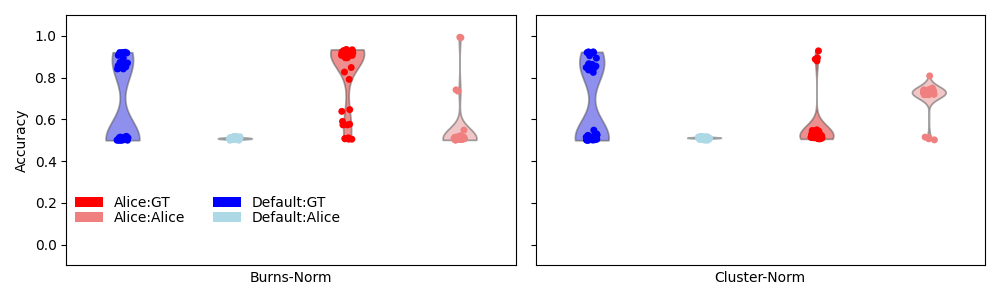}
    \caption{Phi-2 - Compared to Phi-3 and the other models, Phi-2 seems to be an outlier, where probes using Cluster-Norm perform worse than those using Burns-Norms. Possibly, compared to the others, the model is generally less capable.}
    \label{fig:experiment_2_phi_2}
\end{figure*}

\begin{figure*}[htbp]
    \centering
    \includegraphics[width=\linewidth]{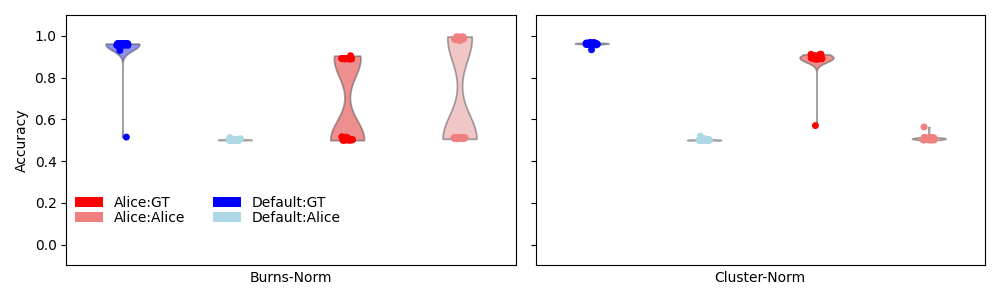}
    \caption{Gemma-7B - Explicit Opinion Experiment}
    \label{fig:experiment_2_gemma}
\end{figure*}

\begin{figure*}[htbp]
    \centering
    \includegraphics[width=\linewidth]{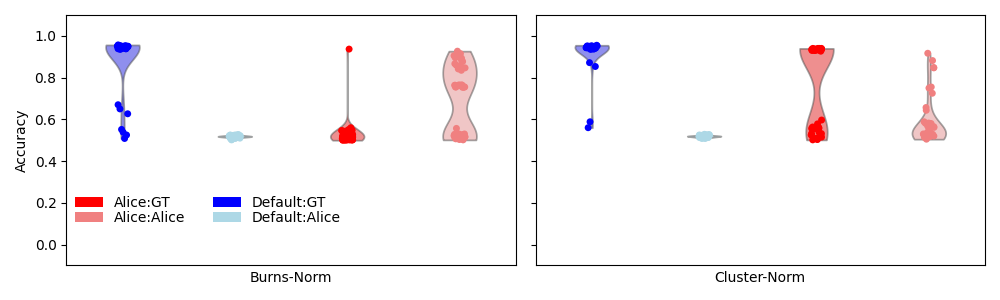}
    \caption{Pythia-6.9B - Explicit Opinion Experiment}
    \label{fig:experiment_2_pythia}
\end{figure*}

Additional violin plots are displayed in the following figures: Llama-3-8B in Figure \ref{fig:experiment_2_llama}, Phi-3 in Figure \ref{fig:experiment_2_phi_3}, Phi-2 in Figure \ref{fig:experiment_2_phi_2}, Gemma-7B in Figure \ref{fig:experiment_2_gemma}, and Pythia-6.9B in Figure \ref{fig:experiment_2_pythia}.

\begin{figure*}[htbp]
    \centering
    \includegraphics[width=\linewidth]{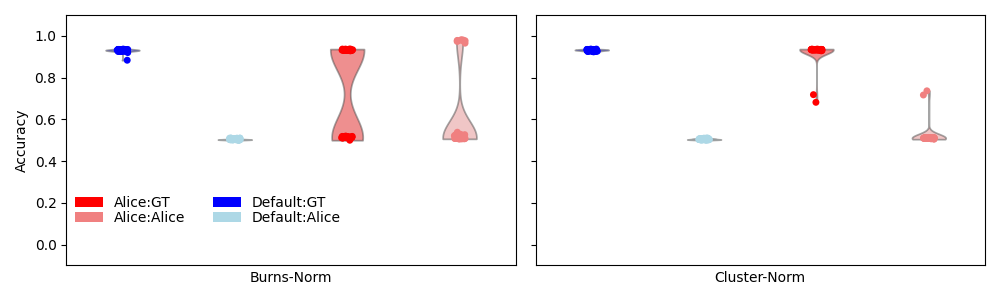}
    \caption{Llama-3-8B - Explicit Opinion Experiment}
    \label{fig:experiment_2_llama}
\end{figure*}

\subsection{Prompt Template Sensitivity}
\label{app:exp3_details}

In \citet{farquhar2023challenges} an analogous experiment investigation prompt template sensitivity is performed using the TruthfulQA \citep{evans2021truthful} dataset. After a manual inspection of this dataset we feel the inclusion of numerous ambiguous questions casts doubt on experimental results, and for this reason we perform the experiments in Section \ref{sec:agent_simulation} using the CommonClaim \citep{casper2023explore} dataset instead. Here, we repeat these experiments using TruthfulQA to allow for a direct comparison to the results in \citet{farquhar2023challenges}.

\begin{figure*}[htbp]
    \centering
    \includegraphics[width=.9\linewidth]{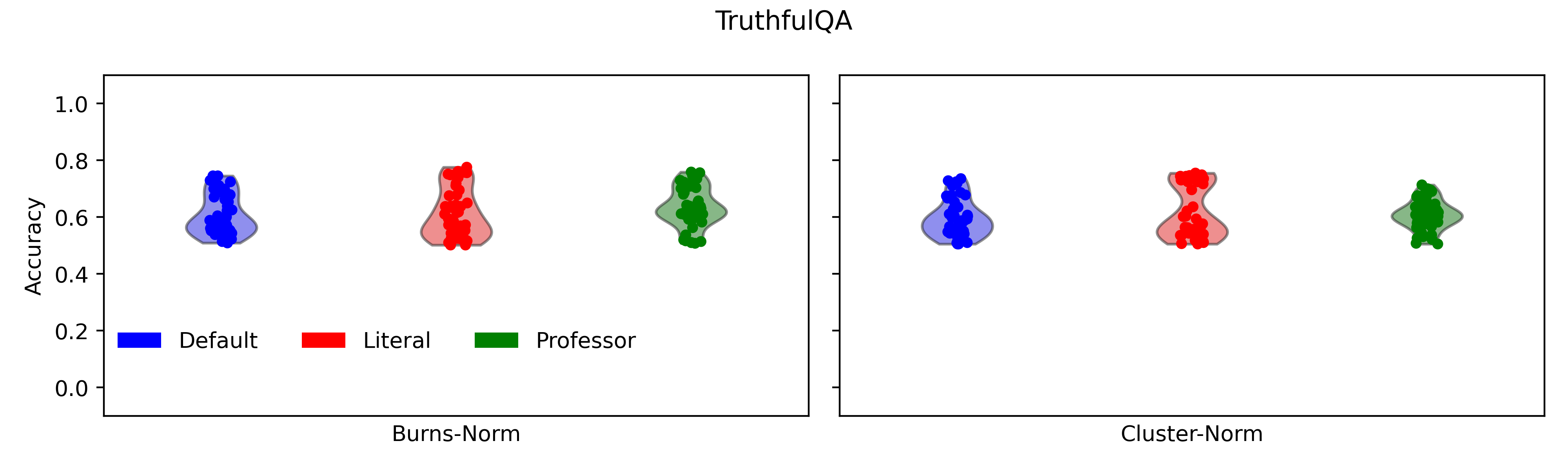}
    \caption{Variation in probe accuracy for the prompt template sensitivity experiment on TruthfulQA for Mistral-7B, at the 75th percentile layer. Contrary to the CommonClaim results (figure \ref{fig:exp4_ccs}), variance is too high to be able to conclude anything.}
    \label{fig:exp4_tqa}
\end{figure*}

Analogous results to those in Figure \ref{fig:exp4_ccs} when performed instead on the TruthfulQA dataset are shown in Figure \ref{fig:exp4_tqa}. We note a high variance in probe accuracy in all settings, and therefore feel these experimental results do not lead to any clear conclusions.

\begin{figure*}[htbp]
    \centering
    \includegraphics[width=\linewidth]{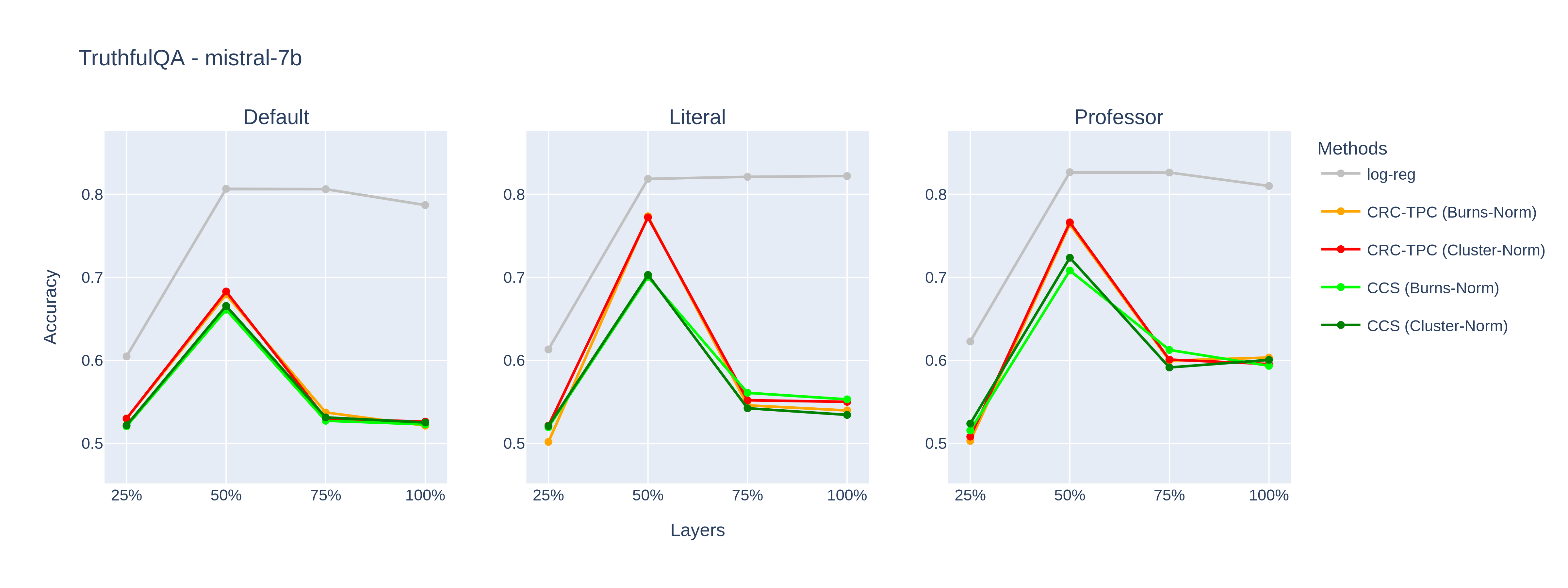}
    \caption{Mean accuracy of Logistic Regression, CRC, and CCS probes over 50 probes, using Mistral-7B, for each prompt template described in Section \ref{sec:agent_simulation}.}
    \label{fig:exp4_mistral_all-layers_tqa}
\end{figure*}

\begin{figure*}[htbp]
    \centering
    \includegraphics[width=\linewidth]{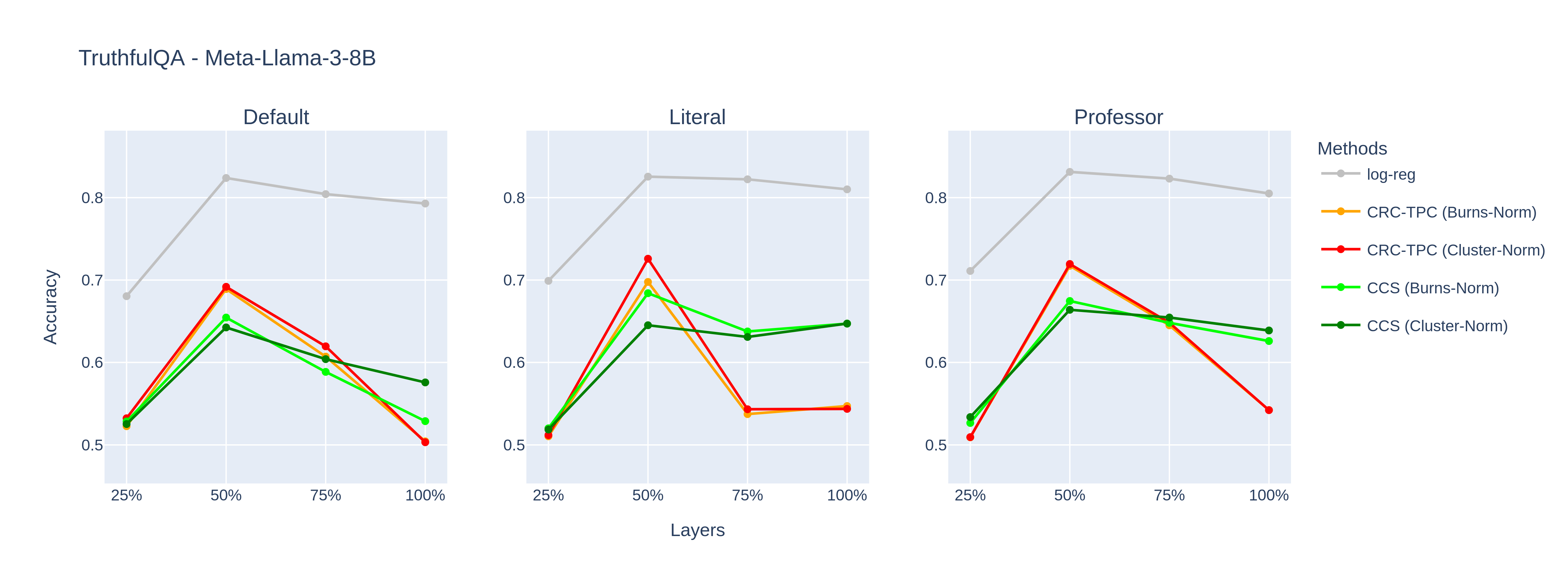}
    \caption{Mean accuracy of Logistic Regression, CRC, and CCS probes over 50 probes, using Llama-3-8B, for each prompt template described in Section \ref{sec:agent_simulation}.}
    \label{fig:exp4_llama_all-layers_tqa}
\end{figure*}

\begin{figure*}[htbp]
    \centering
    \includegraphics[width=\linewidth]{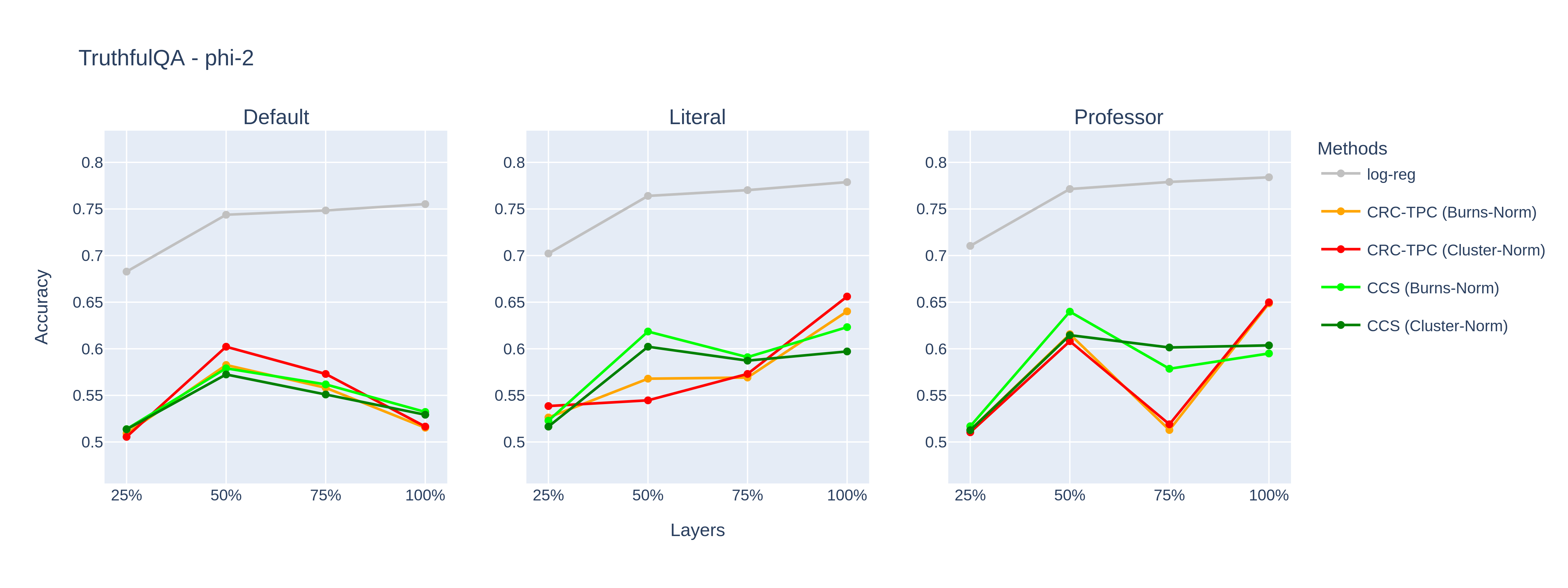}
    \caption{Mean accuracy of Logistic Regression, CRC, and CCS probes over 50 probes, using Phi-2, for each prompt template described in Section \ref{sec:agent_simulation}.}
    \label{fig:exp4_phi_all-layers_tqa}
\end{figure*}

We thoroughly verify these results by repeating these experiments when harvest contrast pair activations at the 25th percentile, 50th percentile, and last layer for Mistral-7B, as well as two additional models: Llama-3-8B and Phi-2. These results are visualized in Figures \ref{fig:exp4_mistral_all-layers_tqa} to \ref{fig:exp4_phi_all-layers_tqa}.

\begin{figure*}[htbp]
    \centering
    \includegraphics[width=\linewidth]{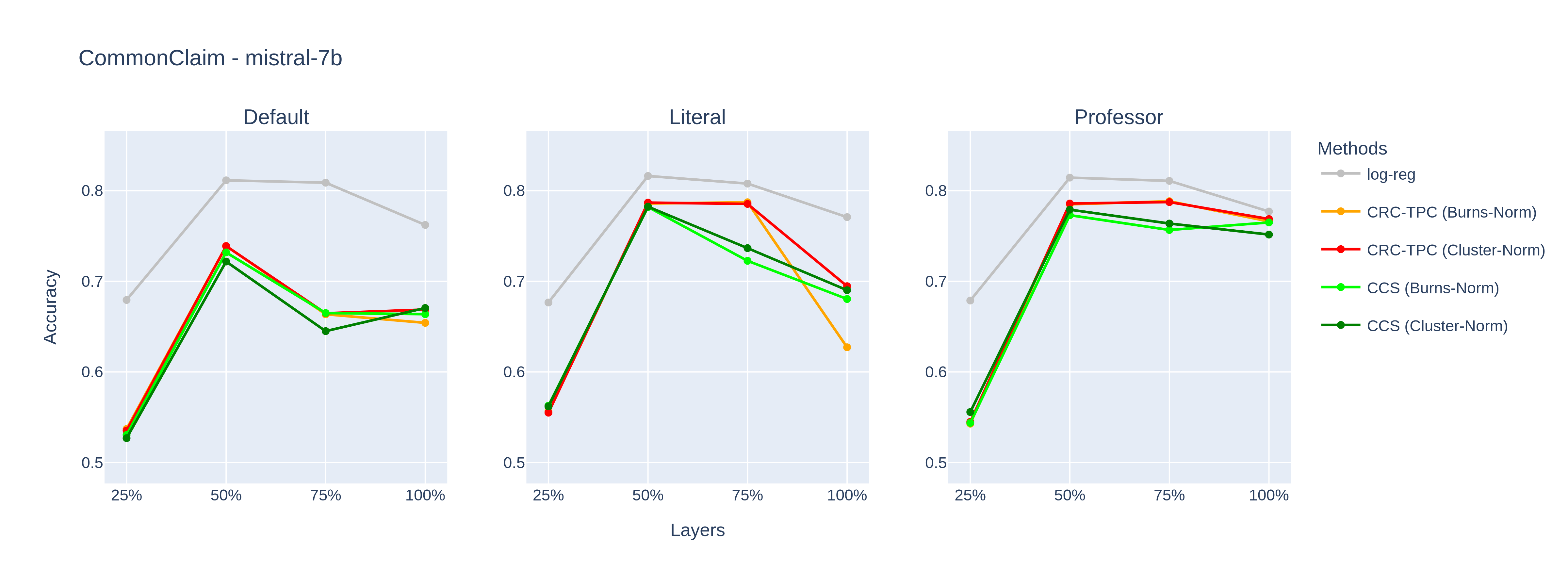}
    \caption{Mean accuracy of Logistic Regression, CRC, and CCS probes over 50 probes, using Mistral-7B, for each prompt template described in Section \ref{sec:agent_simulation}.}
    \label{fig:exp4_mistral_all-layers_cc}
\end{figure*}

\begin{figure*}[htbp]
    \centering
    \includegraphics[width=\linewidth]{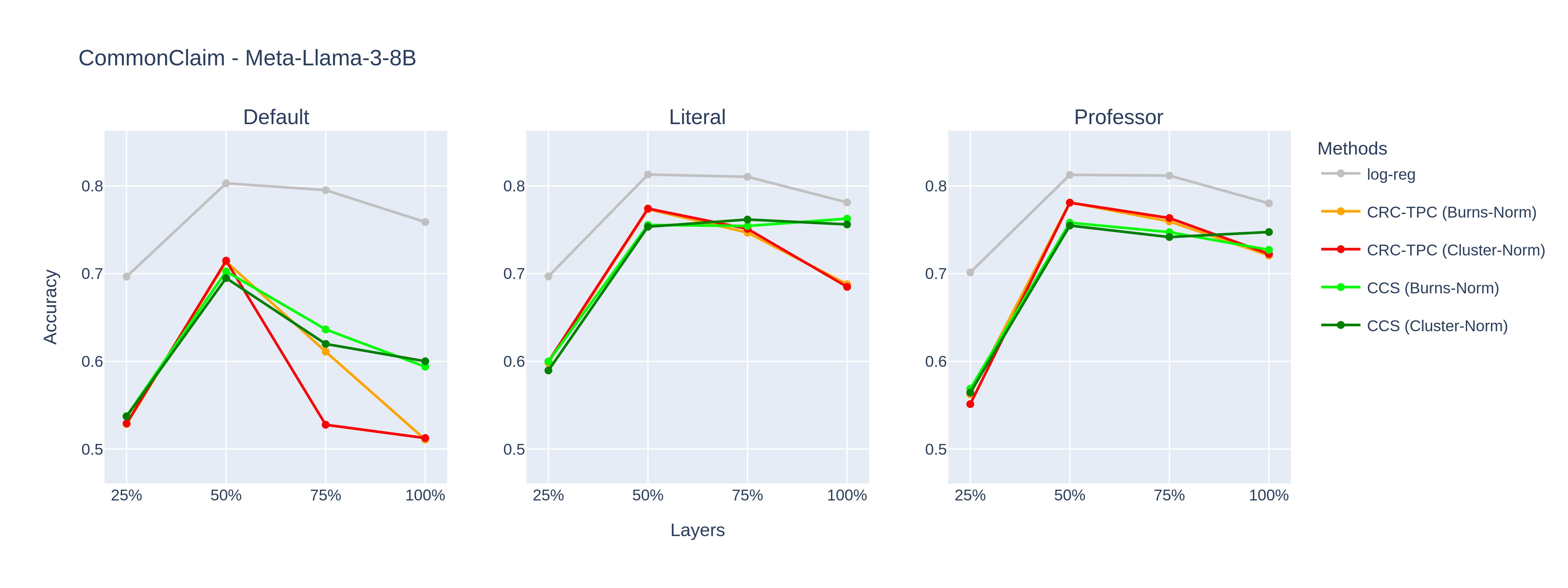}
    \caption{Mean accuracy of Logistic Regression, CRC, and CCS probes over 50 probes, using Llama-3-8B, for each prompt template described in Section \ref{sec:agent_simulation}.}
    \label{fig:exp4_llama_all-layers_cc}
\end{figure*}

\begin{figure*}[htbp]
    \centering
    \includegraphics[width=\linewidth]{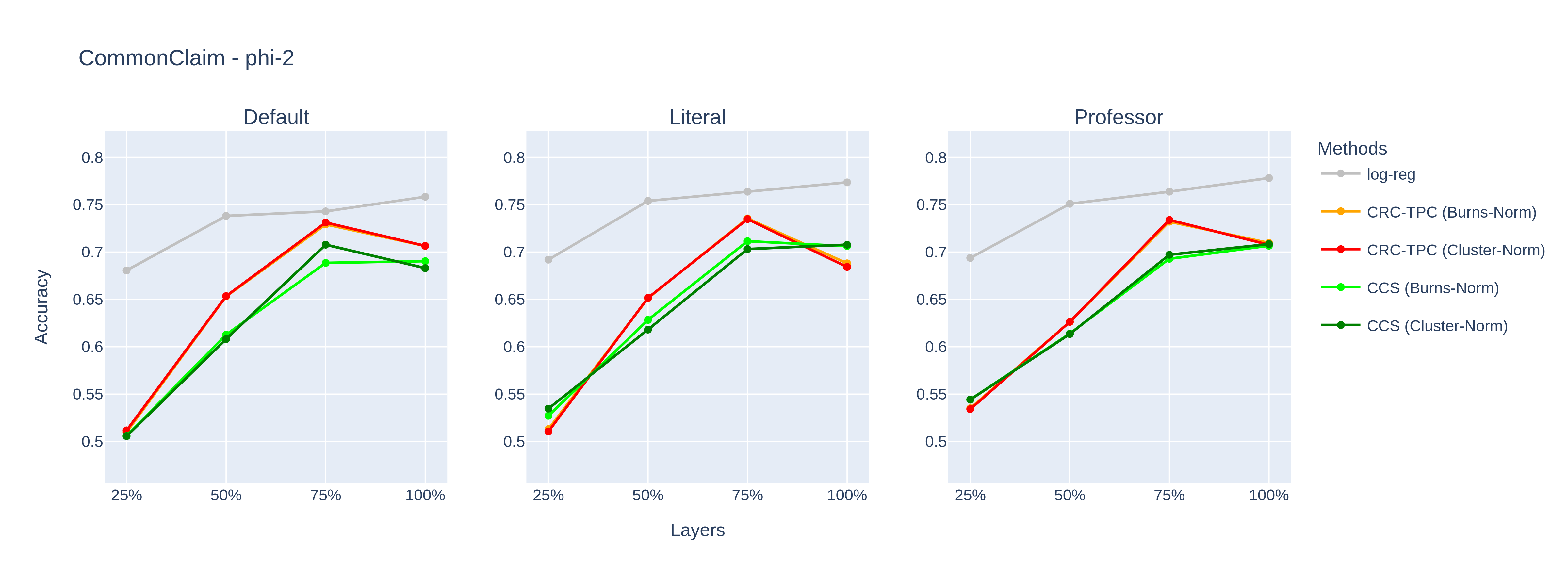}
    \caption{Mean accuracy of Logistic Regression, CRC, and CCS probes over 50 probes, using Phi-2, for each prompt template described in Section \ref{sec:agent_simulation}.}
    \label{fig:exp4_phi_all-layers_cc}
\end{figure*}

We additionally repeat these layer-by-layer experiments, again using the same two additional models, for the experiments outlined in Section \ref{sec:agent_simulation} using the CommonClaim dataset. Results are visualized in Figures \ref{fig:exp4_mistral_all-layers_cc} to \ref{fig:exp4_phi_all-layers_cc}.

\begin{figure*}[htbp]
    \centering
    \includegraphics[width=\linewidth]{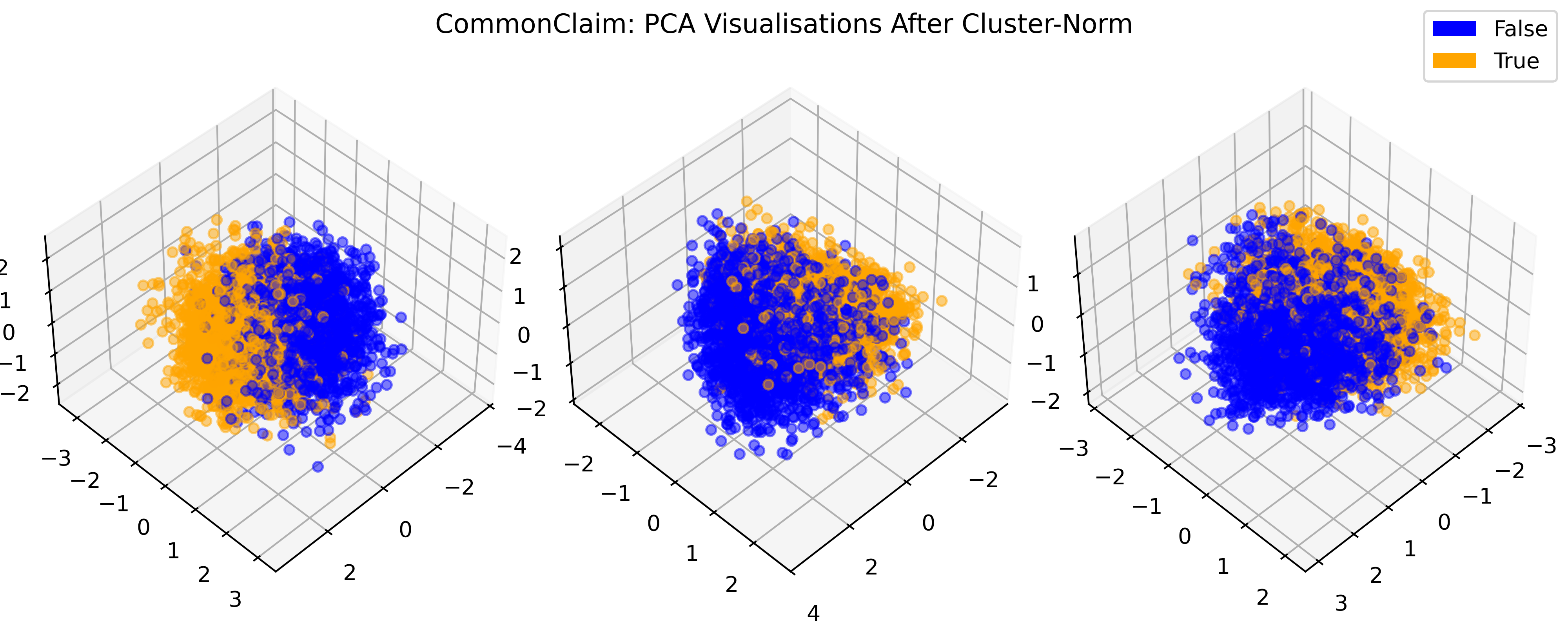}
    \caption{Visualization of the top three PC of $\widetilde{\mathcal{M}}(x_i^+) - \widetilde{\mathcal{M}}(x_i^-)$ - with per cluster normalization - respectively from left to right : for the default, literal and professor prompts. True and False correspond to the ground truth label of these question-answering prompts. We see no notable difference between the three settings, and there is no difference at all to be seen between Burns-Norm and Cluster-Norm. If anything, we can see that in the literal and professor settings, the separation between True and False is slightly more aligned with the first PC.}
    \label{fig:exp4_cc_pca}
\end{figure*}

We present results on TruthfulQA for Mistral-7B in figure \ref{fig:exp4_tqa}, following the exact same procedure as in the main body for CommonClaim (Figure \ref{fig:exp4_ccs}). We then show a PCA visualization of contrast differences for CommonClaim in figure \ref{fig:exp4_cc_pca}.

\section{Zero-Shot Results}

For reference, we present the zero-shot results for different models.

\subsection{Random Words}

For the random word experiment, we report only the zero-shot results using the \textit{default} prompts, where no random word is appended. This decision stems from the structure of our biased prompts and the nature of zero-shot prediction. Recall that for biased prompts we append a random word after the sentiment \textit{positve} or \textit{negative} pseudo-label:

{\small
\begin{verbatim}
Consider the following example: [review],
Between positive and negative, the
sentiment of this example
is [label]. [random_word]
\end{verbatim}
}

However, in the zero-shot scenarios, the model would need to predict the label itself. Consequently, using modified prompts for zero-shot prediction becomes impractical here. We therefore limit our reporting to results from the default prompts for this experiment. The results are presented in Table \ref{table:exp1_zeroshot}. On average, The zero-shot accuracies observed in this experiment are notably lower than the probe acccuracies for the 75th percentile layer, as illustrated in Figure \ref{fig:exp_1_average}. 

\begin{table}[h!]
\centering
\begin{tabular}{l|r}
\hline
\textbf{Model} & \textbf{Default} \\
\hline
Gemma-7B     & 0.51 \\
Llama-3-8B   & 0.76 \\
Mistral-7B   & 0.86 \\
Phi-2        & 0.84 \\
Phi-3        & 0.94 \\
Pythia-6.9B  & 0.51 \\
\hline
\end{tabular}
\caption{Random Word experiment: Zero-shot prompting performance for different models on the IMDb dataset for the \textit{default} prompt templates only (no random word is appended to a prompt).}
\label{table:exp1_zeroshot}
\end{table}

\subsection{Explicit Opinion}
We evaluate zero-shot performance using unbiased (standard) and biased (with added explicit opinion) prompt templates across six language models: Gemma-7B, Llama-3-8B, Mistral-7B, Phi-2, Phi-3, and Pythia-6.9B. Table \ref{table:exp2_zeroshot} presents the results. Similar to the random words experiment, we observe that the zero-shot accuracies are, on average, lower than the probe acccuracies for both default and modified prompt templates, as illustrated in Figure \ref{fig:exp_2_average}. However, the difference between the zero-shot accuracies for default and modified prompts is not substantial for most models.

\begin{table}[h!]
\centering
\begin{tabular}{l|rr}
\hline
\textbf{Model} & \textbf{Default} & \textbf{With Explicit Opinion} \\
\hline
Gemma-7B     & 0.51 & 0.51 \\
Llama-3-8B   & 0.59 & 0.60 \\
Mistral-7B   & 0.68 & 0.59 \\
Phi-2        & 0.54 & 0.68 \\
Phi-3        & 0.93 & 0.92 \\
Pythia-6.9B  & 0.50 & 0.50 \\
\hline
\end{tabular}
\caption{Explicit Opinion experiment: Zero-shot performance comparison on the IMDb dataset. Results show accuracies for default prompt templates (no explicit opinion is added) and modified prompt templates (with added explicit opinion).}
\label{table:exp2_zeroshot}
\end{table}

\subsection{Prompt Template Sensitivity}


\begin{table}[ht]
\centering
\begin{tabular}{l|rrr}
\hline
\multirow{2}{*}{\textbf{Model}} & \multicolumn{3}{c}{\textbf{Prompt Template}} \\
 & \textbf{Default} & \textbf{Literal} & \textbf{Professor} \\
\hline
Mistral-7B   & 0.72 & 0.71 & 0.65 \\
Llama-3-8B   & 0.69 & 0.64 & 0.72 \\
Phi-2        & 0.54 & 0.59 & 0.54 \\
\hline
\end{tabular}
\caption{Zero-shot performance for the Prompt Template Sensitivity experiment.}
\label{table:exp4_zeroshot}
\end{table}

In this experiment, we evaluate zero-shot performance using the same three templates: \textit{Default}, \textit{Literal}, and \textit{Professor}, tested with Mistral-7B, Llama-3-8B, and Phi-2. The corresponding zero-shot accuracies are shown in Table~\ref{table:exp4_zeroshot}. Notably, we do not necessarily see higher performance using the ``Professor'' template for all models. While this outcome is unexpected, it is difficult to speculate on its underlying cause; we instead focus on the effect of cluster normalization in our main discussion of results.

\subsection{Implicit Opinion}

\begin{table}[ht!]
\centering
\begin{tabular}{l|rrrr}
\hline
\multirow{2}{*}{\textbf{Model}} & \multicolumn{2}{c}{\textbf{Company}} & \multicolumn{2}{c}{\textbf{Non-Company}} \\
 & \textbf{Unbiased} & \textbf{Biased} & \textbf{Unbiased} & \textbf{Biased}\\
\hline
Mistral-7B   & 0.96 & 0.39 & 0.98 & 0.62 \\
Llama-3-8B   & 0.97 & 0.17 & 0.98 & 0.53 \\
Phi-2        & 0.98 & 0.39 & 0.94 & 0.90 \\
\hline
\end{tabular}
\caption{Zero-shot performance performance for the Implicit Opinion Experiment. Here, the effect of Alice's biased implicit opinion is clearly demonstrated.}
\label{table:exp3_zeroshot}
\end{table}

Following the setup of this experiment, we examine zero-shot performance for both ``company'' and ``non-company'' questions, tested with Mistral-7B, Llama-3-8B, and Phi-2. Our results are summarized in Table~\ref{table:exp3_zeroshot}.

Focusing on Mistral-7B, but with the same pattern of results apparent for all three models, introducing Alice's biased implicit opinion substantially decreases zero-shot accuracy for questions labeled as ``\textit{company}'', from 0.96 to 0.39. Conversely, for non-company labels, accuracy declines moderately from 0.98 to 0.62 when the biased setting is applied. These results underscore the significant influence of biased prompts on model performance, particularly in scenarios involving implicit opinions.

\end{document}